\documentclass[10pt,twocolumn,letterpaper]{article}

\usepackage{iccv}
\usepackage{times}
\usepackage{epsfig}
\usepackage{graphicx}
\usepackage{amsmath}
\usepackage{amssymb}
\usepackage{amssymb}
\usepackage{booktabs}
\usepackage{multirow}
\usepackage{paralist}
\usepackage{microtype}
\usepackage{soul}
\usepackage{xcolor}
\usepackage{pifont}
\newcommand{\xmark}{\ding{55}}
\usepackage[pagebackref=true,breaklinks=true,letterpaper=true,colorlinks,bookmarks=false,citecolor=blue,linkcolor=blue]{hyperref}

\iccvfinalcopy % *** Uncomment this line for the final submission

\def\iccvPaperID{3494} % *** Enter the ICCV Paper ID here

\ificcvfinal\fi

\begin{document}

\title{Learning Cross-Modal Contrastive Features for Video Domain Adaptation}

\author{Donghyun Kim$^1$, Yi-Hsuan Tsai$^2$, Bingbing Zhuang$^2$, Xiang Yu$^2$ \\
Stan Sclaroff$^1$, Kate Saenko$^{1,3}$, Manmohan Chandraker$^2$ \\
$^1$Boston University, $^2$NEC Labs America, $^{3}$MIT-IBM Watson AI Lab \\
\tt\small $^1$\{donhk, sclaroff, saenko\}@bu.edu, $^2$\{ytsai,  bzhuang, xiangyu, manu\}@nec-labs.com}

\maketitle

%%%%%%%%% ABSTRACT
\begin{abstract}
Learning transferable and domain adaptive feature representations from videos is important for video-relevant tasks such as action recognition.
Existing video domain adaptation methods mainly rely on adversarial feature alignment, which has been derived from the RGB image space.
However, video data is usually associated with multi-modal information, e.g., RGB and optical flow, and thus it remains a challenge to design a better method that considers the cross-modal inputs under the cross-domain adaptation setting.
%
% However, due to the particular characteristic in videos that usually consider the multi-modality property, e.g., RGB and optical flow, it remains a challenge to design a better adaptation method that considers the cross-modal inputs across domains.
% \mc{MC: This sentence is vague .... particular characteristic? challenge?} \mc{MC: You need to set up by this point what is meant by ``cross-modal'' and ``cross-domain''.}
%
To this end, we propose a unified framework for video domain adaptation, which simultaneously regularizes cross-modal and cross-domain feature representations.
Specifically, we treat each modality in a domain as a view and leverage the contrastive learning technique with properly designed sampling strategies.
As a result, our objectives regularize feature spaces, which originally lack the connection across modalities or have less alignment across domains.
% \mc{MC: Need to be more specific on the novel methodological contributions.}
%
We conduct experiments on domain adaptive action recognition benchmark datasets, \ie, UCF, HMDB, and EPIC-Kitchens, and demonstrate the effectiveness of our components against state-of-the-art algorithms.
% \mc{MC: Be specific on what datasets and the extent of improvement.}

\end{abstract}

%%%%%%%%% BODY TEXT

\section{Introduction}

Recently, domain adaptation has gained a lot of attention due to its efficiency during training without the need of collecting ground truth labels in the target domain.
%
% Despite significant progress in domain adaptation across image classification \cite{long2015learning,ganin2016domain,tzeng2017adversarial,Saito_CVPR_2018}, semantic segmentation \cite{Hoffman_ICML_2018,Tsai_DA4Seg_ICCV19,Vu_CVPR_2019,Li_CVPR_2019} and object detection \cite{YChenCVPR18,SaitoCVPR19,KimCVPR19,Hsu_uda_det_eccv20}, most efforts in computer vision are still limited to image data.
% %
% In this paper, we focus on adapting video data, especially for the application of action recognition.
%
Existing methods have made significant progress in image-based tasks, such as classification \cite{long2015learning,ganin2016domain,tzeng2017adversarial,Saito_CVPR_2018}, semantic segmentation \cite{Hoffman_ICML_2018,Tsai_DA4Seg_ICCV19,Vu_CVPR_2019,Li_CVPR_2019,Paul_daseg_eccv20} and object detection \cite{YChenCVPR18,SaitoCVPR19,KimCVPR19,Hsu_uda_det_eccv20}.
While several works have sought to extend this success to video-based tasks like action recognition by aligning appearance (\eg, RGB) features through adversarial learning \cite{Chen_da_iccv19,drone_wacv20,Pan_aaai20}, challenges persist in video adaptation tasks due to the greater complexity of the video data.
% Previous video domain adaptation approaches \cite{drone_wacv20,Chen_da_iccv19} mainly focus on aligning RGB features via adversarial learning \cite{ganin2016domain,tzeng2017adversarial} as used in the image domain.
%
Moreover, different from the image data, domain shifts in videos for action recognition often involve more complicated environments, which increases the difficulty for adaptation.
For example, the “fencing” action usually happens in a stadium, but it can happen in other places
such as home or outdoors. Also, different actions can take place under the same background.
Therefore, purely relying on aligning RGB features can be biased to the background and affect the performance.

\begin{figure}[!t]
	\centering
	\includegraphics[width=0.9\linewidth]{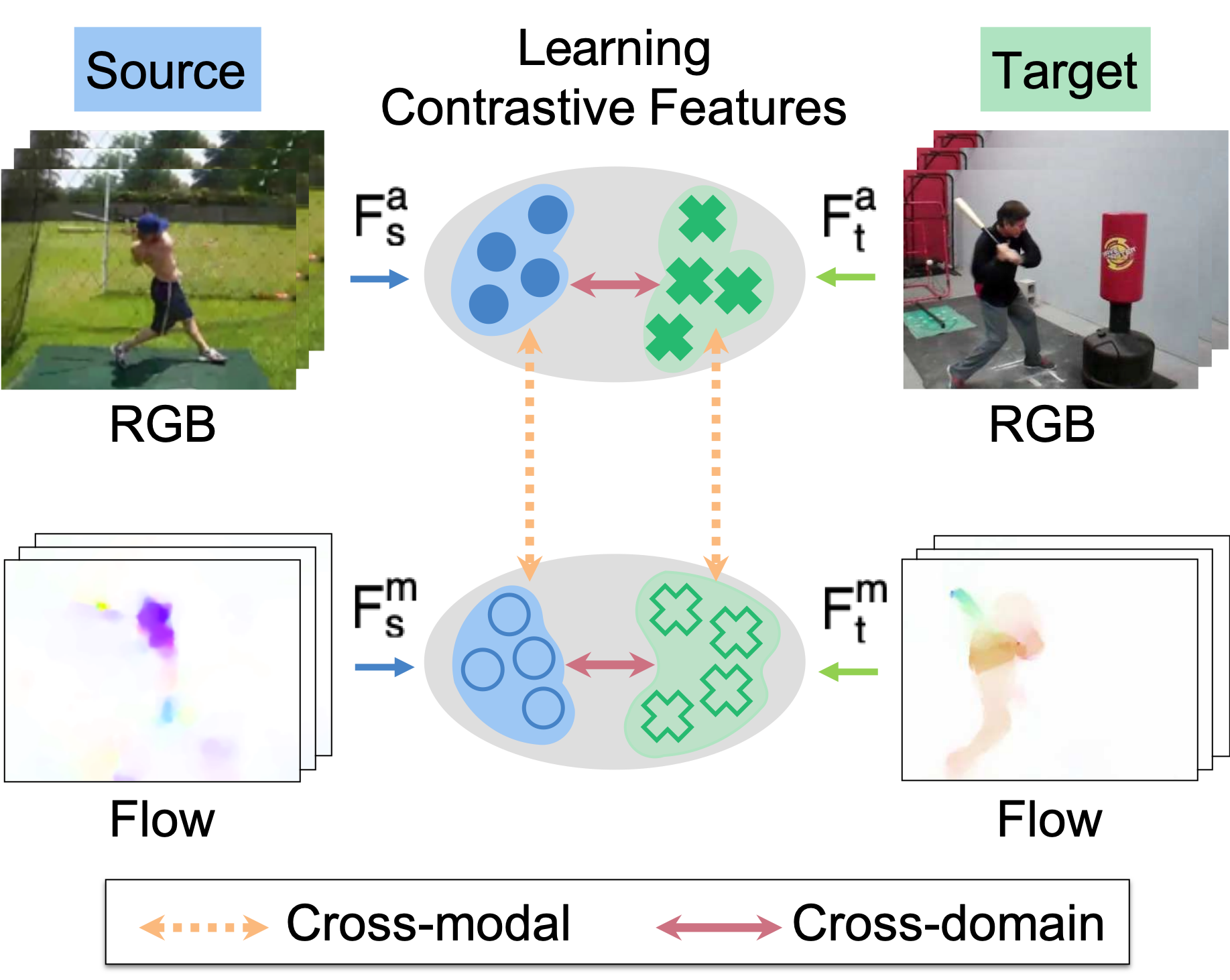}
	%\vspace{-4mm}
	\caption{
	We propose a cross-modal contrastive learning framework for video domain adaptation. Our framework consists of two contrastive learning objectives: (1) cross-modal contrastive learning to align cross-modal representations from the same video, and (2) cross-domain contrastive learning to align representations between the source and target domains in each modality.
	}
	\label{fig:teaser}
	\vspace{-4mm}
\end{figure}

% \mc{MC: The start is awkward, since there are several methods for video adaptation (or broadly unsupervised learning in videos), so it appears as if the first paragraph is not acknowledging them. Rather, you can have the first paragraph go as: Need for domain adaptation. Strong progress in image-based tasks. While several works have sought to extend this success to video-based tasks like action recognition by aligning appearances \cite{}, challenges persist due to the greater complexity of videos. For example (give your example). In this work ….}

In addition to the appearance cue, other modalities such as motion, audio, and text are considered in (self-)supervised learning methods on the video data \cite{Simonyan_nips_14,Arandjelovic_iccv17,Korbar_nips18,Piergiovanni_cvpr20}.
In this work, we focus on appearance and motion as the two most common modalities in the cross-domain action recognition task, in which the motion modality (\ie, optical flow) is shown to be more domain-invariant (\eg, background changes) than RGB \cite{Munro_cvpr20}.
As a result, motion can better capture background-irrelevant information, while RGB can identify semantically meaningful information under different camera setups, e.g., camera perspective.

As shown in Figure \ref{fig:teaser}, with two modalities across two domains, adaptation becomes a task of how to explore the relationship of cross-modal and cross-domain features, to fully exploit the multi-modal property for video domain adaptation.
That is, given either the source video $V_s$ or the target one $V_t$, they can be associated to either the appearance feature $F^a$ or the motion feature $F^m$, which results in four combinations of feature spaces, i.e., $F_s^a$, $F_t^a$, $F_s^m$, $F_t^m$.
Thus, the ensuing task is to design an effective adaptation mechanism for dealing with these four feature spaces.
%
% We note that the prior work \cite{Munro_cvpr20} also adopts a multi-modal framework, but this work focuses on typical adversarial alignment and a self-supervised objective to predict whether the RGB/flow modality comes from the same video clip, without the exploration of the cross-modal and cross-domain features like our work.
% \mc{MC: This is vague unless reader is familiar with Munro et al.~work, be more specific how our exploration is different}
%
% Compared to that work, our aim is different as we would like to explore the relationship of cross-modality and cross-domain features, in order to fully exploit the multi-modality property for video domain adaptation.
%
Since each modality has its characteristics and benefit (e.g., flow is more domain-invariant and RGB can capture semantic cues), it is of great interest to enable feature learning across the two modalities.
% \mc{MC: This is superfluous, rather previous paragraphs should already establish need for cross-modal alignment}
%
Our key contribution stems from the observation that typical adversarial feature alignment schemes used in e.g. \cite{Chen_da_iccv19,Choi_eccv20} may not be directly applied in the cross-modal setting. For example, it is not reasonable to directly align the RGB feature $F_s^a$ in the source domain with the flow feature $F_s^m$ or $F_t^m$ in either domain.

To tackle this issue, motivated by the recent advancements in self-supervised multi-view learning \cite{cmc_eccv20} that achieves powerful feature representations, we propose to treat each modality as a view, while introducing the cross-domain video data in our multi-modal learning framework.
To this end, we leverage the contrastive learning objectives for performing feature regularization mutually among those four feature spaces (see Figure \ref{fig:teaser}) under the video domain adaptation setting.
We note that the prior work \cite{Munro_cvpr20} also adopts a multi-modal framework, but it focuses on typical adversarial alignment and a self-supervised objective to predict whether the RGB/flow modality comes from the same video clip, without the exploration of jointly regularizing cross-modal and cross-domain features like our work.

% \mc{MC: Maybe elaborate a bit differently so it doesn't convey incorrectly that we only apply CMC.} 
% , achieving good empirical performance.

%\BB{Seems it is not clear to see a strong motivation of using contrastive learning. If $F_s^a$ cannot be directly aligned to $F_s^m$ or $F_t^m$, they can also use a projection head $h$ as we do to solve the problem, without using contrastive learning? "regularization" is weaker than "alignment" so it has no above problem? Can we elaborate more on why contrastive learning is truly needed?}

%
More specifically, our framework is allowed to contrast features across modalities within a domain (e.g., between $F_s^a$ and $F_s^m$) or across domains using one modality (e.g., between $F_s^a$ and $F_t^a$). Two kinds of loss functions are designed accordingly: 1) a cross-modal loss that considers each modality as one view in a video while contrasting views in other videos from the same domain; 2) a cross-domain loss that considers one modality at a time and contrasts features based on the (pseudo) class labels of videos across two domains.
% 1) a cross-modality loss that considers each modality as one view, where we contrast video features (for both source and target domains) based on whether the feature is extracted from the same video, e.g., within the same video, one positive pair could be $F^a$ and $F^m$; and 2) a cross-domain loss that contrasts features in each modality but from different domains.
% %
% Since we do not have the action labels in the target domain, we obtain the pseudo labels in an online fashion and construct positive/negative samples for the target videos across domains.
%
%Besides achieving good performance, 
There are several benefits of the proposed contrastive learning-based feature regularization strategies: 1) it is a unified framework that allows the interplay across features in different modalities and domains, while still enjoying the benefits of each modality; 2) it enables sampling strategies of selecting multiple positive and negative samples in the loss terms, coupled with memory banks to record large variations in video clips; 3) our cross-domain loss can be considered as a soft version of pseudo-label self-training with the awareness of class labels, which performs more robustly than typical adaptation methods.
% \mc{MC: There can be some mention of important observations with respect to design choices such as sampling strategy and memory banks.}
% As a result, our method allows the interplay across features in different modalities and domains via our proposed two feature regularization strategies, while still enjoying the benefits of each modality.
% \YH{We may need to mention why contrastive learning is useful here, e.g., allow a unified framework, multiple positive/negative samples with memory banks, a soft version of pseudo-label self-training.}
% where such feature regularization can mitigate the modality/domain gap problem and hence improve the domain adaptive action recognition accuracy.
%

We conduct experiments in video action recognition benchmark datasets, including the UCF \cite{ucf} $\leftrightarrow$ HMDB \cite{hmdb} setting, and the EPIC-Kitchens \cite{epic-kitchen,Munro_cvpr20} dataset.
We show that including either our cross-modal or cross-domain contrastive learning objective improves accuracy while combining these two strategies in a unified framework obtains the best results.
Moreover, our method performs favorably against state-of-the-art domain adaptation techniques, e.g., adversarial feature alignment \cite{Chen_da_iccv19,Munro_cvpr20}, self-learning scheme \cite{Choi_eccv20}, and pseudo-label self-training.
The main contributions of this work are summarized as follows.
\begin{itemize}
    \item We propose a new multi-modal framework for video domain adaptation that leverages the property in four different feature spaces across modalities and domains.
    
    \item We leverage the contrastive learning technique with well-designed sampling strategies and demonstrate the application to adaptation for cross-domain action recognition by exploiting appearance and flow modalities.
    
    % \item In application to video domain adaptation, our framework regularizes the feature space of appearance and flow modalities across domains by leveraging the contrastive learning technique.
    
    \item We show the effectiveness of both the cross-modal and cross-domain contrastive objectives, by achieving state-of-the-art results on UCF-HMDB and EPIC-Kitchens adaptation benchmarks with extensive analysis. 
    % which provides a new perspective for considering the problem of video domain adaptation.
    
\end{itemize}

\section{Related Work}
In this section, we discuss existing fully-supervised and domain adaptation methods for action recognition, as well as methods on unsupervised learning for video representations.

\vspace{-4mm}
\paragraph{Supervised Action Recognition.}
Action recognition is one of the important tasks for understanding video representations. With the recent advancements in deep learning, early works either adopt 2D \cite{Karpathy_cvpr14} or 3D \cite{Ji_pami13} convolutional networks on RGB video frames, which achieve significant progress.
To improve upon the single-modal framework, optical flow is commonly used as the temporal cue to greatly improve the action recognition accuracy \cite{Simonyan_nips_14}.
Following this multi-modal pipeline, several methods are proposed to further incorporate the long-term temporal context \cite{non-local,slowfast} and structure \cite{Tran_cvpr17,Zhou_eccv18,Wang_eccv16}, or extend to the 3D convolutional networks \cite{Carreira_cvpr17,Tran_iccv15}.
Moreover, recent approaches show the benefit of adopting 1D/2D separable convolutional networks \cite{Tran_cvpr17,Xie_eccv18}, while other methods \cite{slowfast,Jiang_iccv19} focus on improving the 3D convolutional architecture for action recognition, to be computationally efficient.
Despite the promising performance of these methods in a fully-supervised manner, our focus is to develop an effective action recognition framework under the unsupervised domain adaptation setting.

\begin{figure*}[!t]
	\centering
	\vspace{-0.4cm}
	\includegraphics[width=1.0\linewidth]{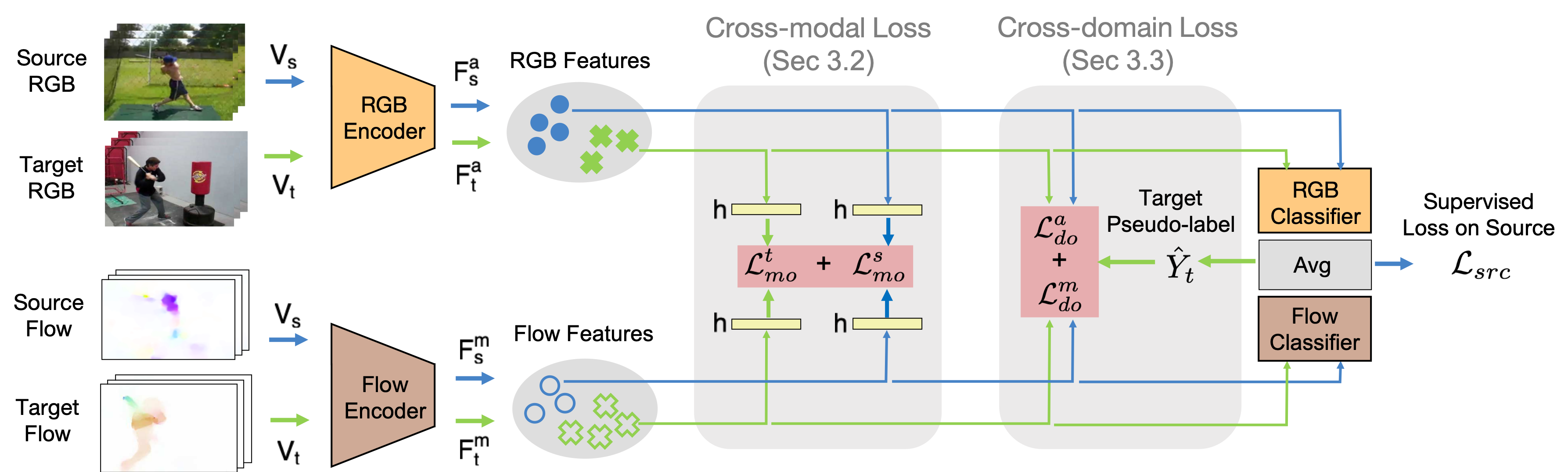}
	\caption{
	An overview of our cross-modal contrastive learning framework. We use a two-stream network for RGB and flow. Each stream takes video clips and outputs feature vectors for each domain and modality ($F_s^a, F_t^a, F_s^m$, $F_t^m$). For cross-modal contrastive learning, we add the projection head ($h$) to learn an embedding where the flow and RGB features from the same video are matched (\eg, $h(F_{s_i}^a)$, $h(F_{s_i}^m)$). For cross-domain contrastive learning, we match the cross-domain features of the same class without the projection head (\eg $F_{s_i}^a$, $F_{t_j}^a$) in the same modality. For the unlabeled target domain, we use high-confidence pseudo-labels $\hat{Y}_t$ to find positive samples in the source domain. 
	}
	\label{fig:overview}
	\vspace{-4mm}
\end{figure*}

\vspace{-4mm}
\paragraph{Domain Adaptation for Action Recognition.}
Due to the convenience of recording videos under various conditions, there is an increasing demand for developing approaches for cross-domain action recognition.
Previous methods focus on the setting of cross-domain transfer learning \cite{Sultani_cvpr14,Zhang_aaai19,Xu_ivc16} or tackle the problem of view-point variance in videos \cite{Kong_tip17,Li_nips18,Rahmani_cvpr15,Sigurdsson_cvpr18}.
However, unsupervised domain adaptation (UDA) for action recognition has received less attention until recently.
Early attempts align distributions across the source and target domains using hand-crafted features \cite{Cao_cvpr10,Jamal_bmvc13}, while recent deep learning based methods \cite{Jamal_bmvc18,Pan_aaai20,drone_wacv20,Chen_da_iccv19,Choi_eccv20,Munro_cvpr20} leverage the insight from UDA on image classification and extend it to the video case.
For instance, approaches \cite{Chen_da_iccv19,Pan_aaai20} utilize adversarial feature alignment \cite{ganin2016domain,tzeng2017adversarial} and propose a temporal version with attention modules. 
Moreover, self-supervised learning strategies are adopted by considering the video properties, such as clip orders \cite{Choi_eccv20}, sequential domain prediction \cite{Chen_cvpr20}, and modality correspondence prediction \cite{Munro_cvpr20} in videos.
Similar to \cite{Munro_cvpr20}, our method also considers the multi-modal property, but focuses on a different problem regime. To be specific, we propose a contrastive learning framework that can better exploit the multi-modality to regularize the feature spaces simultaneously across modalities and across domains, which is previously unstudied.

\begin{table}[t]
\centering
\small
\label{tab:notation}
\caption{Summary of notations.}
\begin{tabular}{l|c}
\toprule
Notation & Meaning\\
\hline
$\{V_s, V_t\}$ & $\{$Source, Target$\}$ video clips \\
% $F^a(\cdot)$ & RGB feature extractor \\
% $F^m(\cdot)$ & Flow feature  extractor \\
$F_{s}^{a}$ & Source appearance feature\\
$F_{s}^{m}$ & Source motion feature \\ 
$F_{t}^{a}$ & Target appearance feature\\
$F_{t}^{m}$ & Target motion feature \\ 
$h(\cdot)$ & Shared projection head \\
$\hat{Y_t}$ & Target pseudo-label \\
\bottomrule
\end{tabular}
\vspace{-4mm}
\end{table}

\vspace{-4mm}
\paragraph{Self-supervised Learning for Video Representation.}
Learning from unlabeled videos is beneficial for video representations as video labeling is expensive. For instance, numerous approaches are developed via exploiting the temporal structure in videos \cite{Fernando_cvpr17,Wei_cvpr18}, e.g., temporal order verification \cite{Misra_eccv16} and sorting sequences \cite{Lee_iccv17}.
By leveraging the temporal connection across frames, patch tracking over time \cite{Wang_iccv15} or prediction of future frames \cite{Srivastava_icml15} also facilitates feature learning in videos.
Moreover, to incorporate the multi-modal information into learning, RGB frames, audio, and optical flow are used to align with each other for self-learning \cite{Arandjelovic_iccv17,Korbar_nips18,Piergiovanni_cvpr20,Han_nips20,alwassel2019self,morgado2020learning,piergiovanni2020evolving}. 
After the learning process, such methods are usually served as a pre-training step for the downstream tasks.
In this paper, we study the UDA setting for cross-domain and cross-modal action recognition, which involves a labeled source dataset and unlabeled target videos.

\section{Proposed Method}

In this section, we first introduce the overall pipeline of the proposed approach for video domain adaptation. Then we describe  individual modules for cross-modal and cross-domain feature regularization, followed by the complete objective in a unified framework using contrastive learning.

\subsection{Algorithm Overview}
Given the source dataset that contains videos $\{s_i\} \in V_{s}$ with its action label set $Y_{s}$, our goal is to learn an action recognition model that is able to perform reasonably well on the unlabeled target video set $\{t_i\} \in V_{t}$.
Since we aim to investigate an effective way to construct a domain adaptive model that leverages the benefit of multi-modal information (i.e., RGB and flow) across domains, we utilize a two-stream network \cite{Munro_cvpr20} that takes the RGB and flow images as the input.
As a result, the two-stream network would output the RGB modality feature $F^a$ and the flow modality feature $F^m$, which forms four different feature spaces across the modality and domain, i.e., $F_s^a$, $F_t^a$, $F_s^m$, $F_t^m$.

In our contrastive learning framework, we jointly consider the relationship of these four spaces via two kinds of contrastive loss functions to regularize features as shown in Figure~\ref{fig:teaser}. First, we treat each modality as a view, extract the RGB/flow features from the same domain (either source or target), and contrast them based on whether the features come from the same video, e.g., the cross-modal features of one video, $F_s^a$ and $F_s^m$, should be closer to each other in an embedding space than others extracted from different video clips.
%  \xy{To illustrate the two modality (rgb and flow) and two domains (source and target), we can simply put a $2\times2$ table here, indicating modality versus domain.}
%
Second, for features across the domains, e.g., $F_s^a$ and $F_t^a$, but within the same modality, we contrast them based on whether the videos are likely to share the same action label. To this end, we calculate the pseudo-labels $\hat{Y}_{t}$ on target videos and form positive/negative samples to perform contrastive learning.
Figure \ref{fig:overview} illustrates our overall framework and Table~\ref{tab:notation} summarizes the notations.
%The overall framework of our method is illustrated in Figure \ref{fig:overview} 
% \xy{It would be good to put a paragraph to summarize all the notations appeared in Figure \ref{fig:overview}, i.e., $F_{s}^{a}$: source domain appearance feature, $F_{s}^{m}$: source domain motion feature, etc...}

% 	\BB{Make this figure smaller to save space. It has some overlap with Fig.2. Say, make the "xxx Encoder smaller", put text "Different video" in between $S_1$ and $S_2$ rather than on top of them}}

\subsection{Cross-modal Regularization}
\label{sec:mo}
Motivated by the unsupervised multi-view feature learning method \cite{cmc_eccv20}, we treat each modality as a view and form positive training samples within the same video, as well as negative samples from different videos. However, the difference is that we cannot directly apply negative pairing to all the videos, as in our problem, the videos from two different domains under the same view could still be largely different because of the domain gap.
% However, our task has two domains and the strategy used in \cite{cmc_eccv20} cannot be directly applied in our method by considering all the videos together.
% \xy{Following multi-view feature learning method, we similarly treat each modality as a view, and draft positive training pairs within the same video. The difference is that we cannot directly incorporate all the videos as training, as in our problem, there are data from two different domains.}
%
% Intuitively, the reason is that even videos from two domains have the same view, their resultant feature spaces (e.g., $F_s^a$ and $F_t^a$) can be very different due to the domain gap.
% \xy{It is straight that data from two different domains under the same view could still be largely different because of the domain gap.}
%
Therefore, it would not be proper to mix source and target videos, and instead, we form a contrastive objective in each domain separately.

\begin{figure}[!t]
	\centering
	\vspace{-0.2cm}
	\includegraphics[width=0.9\linewidth]{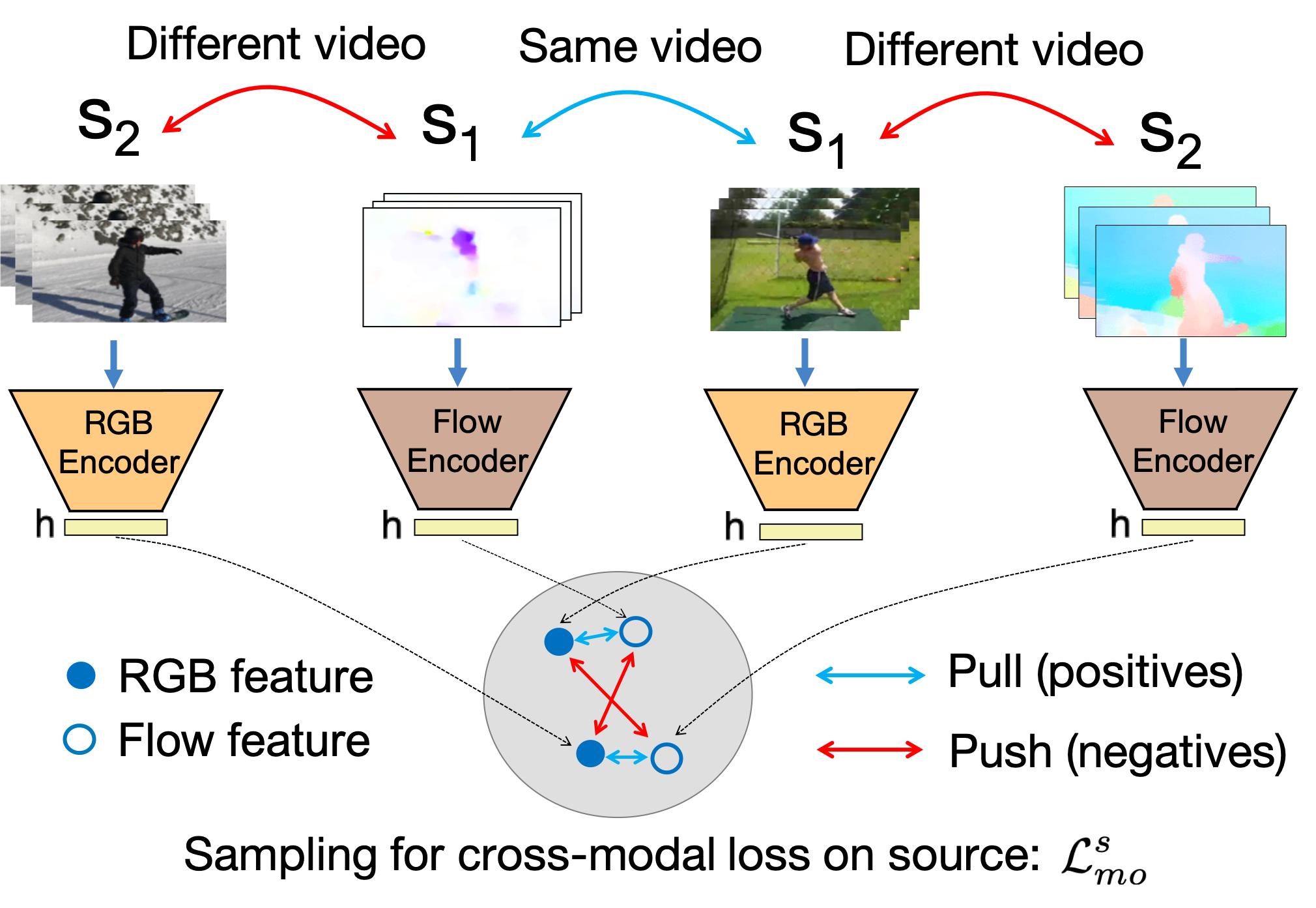}
	\caption{An overview of cross-modal contrastive learning. We pull an RGB feature and a flow feature from the same video clip as positives but push cross-modal features from different video clips.
% 	\YH{TODO, figure smaller}
	}
	\label{fig:sampling_mo}
	\vspace{-4mm}
\end{figure}

\vspace{-4mm}
\paragraph{Sampling Strategy.}
It has been studied that the sampling strategy is crucial in image-based contrastive learning \cite{Kalantidis_nips_20}.
Considering videos from one domain in our case, we select positive training samples from the same video but with different modalities, while sampling the negative ones when the RGB and flow frames are from different videos, regardless of their action labels. An illustration of cross-modal sampling for the source domain is in Figure \ref{fig:sampling_mo}, and a similar strategy is used for the target domain.
% the RGB and flow frames from the same video as a positive training sample, while treating others as negative samples, regardless of their labels.
%

In addition, since one video clip contains many frames, every time we need to randomly sample a window of consecutive frames within a video clip, following the setting in \cite{Choi_eccv20,Munro_cvpr20}.
To account for the large intra-variation within a video clip, we do not assume that the RGB and flow modality need to have the same window of frames. For example, given a video clip, the RGB frames can be randomly sampled from the time window $t\sim t+15$, while the flow frames can be different, e.g., $t+5\sim t+20$. Empirically, we find that such a sampling strategy is especially beneficial for our contrastive learning objective, which is aware of the variation within video clips in the embedding space.
% \xy{I would suggest to put ``Sampling Strategy'' ahead of loss function. In sampling strategy, mainly discuss how the ``posotive'' and ``negative'' paris are sampled.}

% We randomly sample RGB and flow frames in each video clip. We consider only the RGB and flow frames from the same video as a positive sample but treat other as negative samples, regardless of their labels.

% \YH{TODO}
% \BB{Yes, we use different sampling strategies for cross-modality and cross-domain, so better to describe it explicitly rather than letting reviewers figure out themselves from equations. Better put this before Eq.(2).}

\vspace{-4mm}
\paragraph{Similarity Between Samples.}
Another important aspect in contrastive learning is the feature similarity.
Taking the source domain as one example, we have features from RGB and flow, \ie, $F_s^a$ and $F_s^m$.
% Considering the example for the source domain, we have features of $F_s^a$ and $F_s^m$. Note that, each modality maintains its own feature characteristic and they can be complementary to each other, especially for the action recognition task \cite{Simonyan_nips_14}.
%
Since each modality maintains its own feature characteristic, directly contrasting these two features may make the negative impact on the feature representation and reduce the recognition accuracy. To this end, given source features $F_{s_i}$ and $F_{s_j}$ from two videos $\{s_i, s_j\} \in V_s$, we apply an additional projection head $h(\cdot)$ in a way similar to SimCLR \cite{simclr}, and then we can define the similarity function $\phi^s(\cdot)$ between samples with a temperature parameter $\tau$ as:
\begin{equation}
    \label{eq:sim_s}
    \phi^s(F_{s_i}^k, F_{s_j}^l)_{k,l \in \{a, m\}}  = \text{exp}( h(F_{s_i}^k)^\top h(F_{s_j}^l) / \tau ).
\end{equation}
where $\{a, m\}$ represents either the appearance (RGB) or motion (flow) modality.

\vspace{-4mm}
\paragraph{Loss Function.}
Based on the aforementioned sampling strategy and similarity measurement as depicted in Figure \ref{fig:sampling_mo}, the loss function for the source domain is written as:
% We depict the way of sampling positives/negatives on source as one example, and then map them to an embedding space in Figure \ref{fig:sampling_mo}, while a similar way is also applied in the target domain separately.
% Based on this shared embedding space through $h(\cdot)$, the loss function is written as:
%
\begin{equation}
    \label{eq:mo}
    \mathcal{L}_{mo}^s = - \text{log} \frac{\sum\limits_{s_i \in V_s} \phi^s(F_{s_i}^k, F_{s_i+}^l)_{k\neq l} }{ \sum\limits_{s_i \in V_s} \phi^s(F_{s_i}^k, F_{s_i+}^l)_{k\neq l} + \phi^s(F_{s_i}^k, F_{s_j-}^l)_{k\neq l}},
    % \mathcal{L}_{cross-mo}^s = - \text{log} \frac{\sum\limits_{k,l \in \{a, m\}} e^{(h_s(F_{s_i}^k)^\top h_s(F_{s_j}^l) / \tau)}}{},
\end{equation}
% \begin{equation}
%     \label{eq:mo}
%     \mathcal{L}_{mo}^s = - \text{log} \frac{\sum\limits_{k\neq l \in \{a, m\}} \phi^s(F_{s_i}^k, F_{s_i+}^l) }{ \sum\limits_{k\neq l \in \{a, m\}} \phi^s(F_{s_i}^k, F_{s_i+}^l) + \phi^s(F_{s_i}^k, F_{s_j-}^l)},
%     % \mathcal{L}_{cross-mo}^s = - \text{log} \frac{\sum\limits_{k,l \in \{a, m\}} e^{(h_s(F_{s_i}^k)^\top h_s(F_{s_j}^l) / \tau)}}{},
% \end{equation}
%\BB{Looks like Eq.(1) can be much simpler. But first, $F_{s_i+}^l$ and $F_{s_i-}^l$ are not explained in the text, though readers can guess. Then, $\phi^s_+$ and $\phi^s_-$ may cause the confusion that they are different, but they are actually the same function, so using $\phi^s$ is enough, as $F_{s_i+}^l$ and $F_{s_i-}^l$ already indicate positive and negative pairs. Finally, cann't we avoid k,l and just use e.g. $F_{s_i}^a$ and $F_{s_i+}^m$ etc.?}
% \xy{I think this positive and negative samples worth independent illustration. And then in this loss function, we donot need to touch base the notations.}
where $F_{s_i+}$ is the positive sample with a different view (modality) from the same video clip $F_{s_i}$, while $F_{s_j-}$ is the negative sample with another view of a different video from $F_{s_i}$, regardless of their action labels. Here, we omit the notation $k,l \in \{a, m\}$ to have a concise presentation.
% and $F_{s_j-}$ are positive and negative samples for $F_{s_i}$, respectively.
% $\phi^s$ represent the similarity measurement between the features $F_{s_i}$ and $F_{s_j}$ with a temperature parameter $\tau$:
% where $\phi_+^s$ and $\phi_-^s$ represent the similarity measurement for positive/negative pairs between the features $F_{s_i}$ and $F_{s_j}$ with a temperature parameter $\tau$:
% \begin{equation}
%     \label{eq:sim_s}
%     \phi^s(F_{s_i}^k, F_{s_j}^l)_{k\neq l \in \{a, m\}}  = \text{exp}( h(F_{s_i}^k)^\top h(F_{s_j}^l) / \tau ).
% \end{equation}
%
% Note that, to learn cross-modality correspondences in \eqref{eq:mo}, positive samples are only selected from different modalities.
%
\begin{figure}[!t]
	\centering
	\vspace{-0.2cm}
	\includegraphics[width=0.9\linewidth]{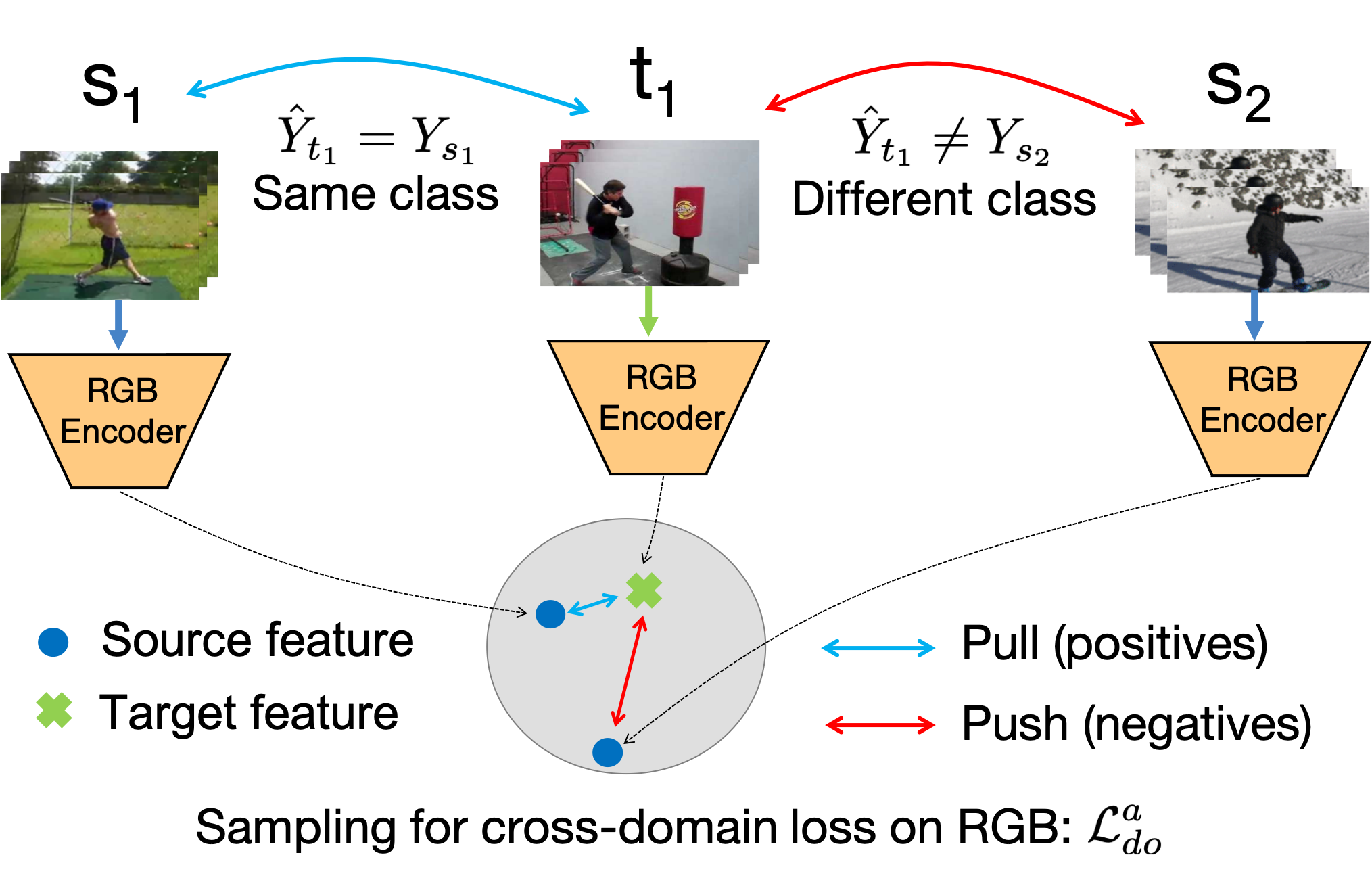}
	\caption{Based on pseudo-labels, we pull source and target features sharing the same labels but push cross-domain features otherwise.
% 	\YH{TODO,  make figure smaller, make embedding shape differently depending on the class}
	}
	\label{fig:sampling_do}
	\vspace{-4mm}
\end{figure}

On the other hand, for videos in the target domain, we construct another loss $\mathcal{L}_{mo}^t$ similar to \eqref{eq:mo} with the same projection head $h(\cdot)$, where the similarity measurement $\phi^t$ between features of target videos $\{t_i, t_j\} \in V_t$ is defined as:
\begin{equation}
    \label{eq:sim_t}
    \phi^t(F_{t_i}^k, F_{t_j}^l)_{k,l \in \{a, m\}} = \text{exp}( h(F_{t_i}^k)^\top h(F_{t_j}^l) / \tau ).
\end{equation}
%
% \xy{the target domain $\tau$ should be different from the source domain $\tau$?}
Here, to consider individual domain characteristics, we find that the key is to form the cross-modal loss for each domain separately, while these two loss functions can still share the same projection head $h$\footnote{We empirically find that using two individual projection heads for each domain produces a similar performance to the one that share the same projection head, so we use the shared projection head as a way for analyzing this embedding space later.}.  The projection head helps to prevent overfitting to this regularization. Without it, the RGB and flow features would be aligned to be the same, so that they are not complementary to each other anymore.
Therefore, by combining these two loss functions in each of the source and target domains (i.e., $\mathcal{L}_{mo}^s$ and $\mathcal{L}_{mo}^t$), features within the same video but from different modalities are closer in an embedding space, which is also served as a feature regularization on unlabeled target videos.
% \xy{is projection head some standard saying? Not sure if embedding layer could be better.}

\subsection{Cross-domain Regularization}
\label{sec:cross_domain}
In addition to the cross-modal regularization introduced in the previous section, we find that there is a missing connection between features across domains.
% because the use of cross-domain regularization in \eqref{eq:sim_s} and \eqref{eq:sim_t} does not involve domain-aligned features.
%
To further exploit the interplay between four feature spaces ($F_s^a$, $F_t^a$, $F_s^m$, $F_t^m$), we propose to use another contrastive learning objective for cross-domain samples.
\vspace{-4mm}
\paragraph{Sampling Strategy via Pseudo-labels.}
Taking one modality, RGB, as the example, we consider cross-domain features of $F_s^a$ and $F_t^a$ in a similar contrastive learning setup as described in Section \ref{sec:mo}. Here, an intuitive way to form positive samples is to find the videos with the same label across domains.
However, since we do not know the action label in the target domain, we first apply our two-stream action recognition model and obtain the prediction score of the target-domain video.
Then, if the score is larger than a certain threshold $T$ (e.g., $T=0.8$ in our setting), we obtain the pseudo-label of this target sample and sample other source videos with the same action label from positive samples (otherwise they are negative samples).
The procedure for the RGB modality is illustrated in Figure~\ref{fig:sampling_do}, while a similar way is used for the flow modality.
% However, since we do not know the action label in the target domain, we online generate pseudo labels $\hat{Y_t}$ based on the prediction score from the source classifier by a threshold $T$ (\ie, $T = 0.8$ in this paper), and then we select samples with the same label in source videos as positive ones and otherwise they are negative samples as shown in Figure~\ref{fig:sampling_do}.
% \xy{again, formally put a separate paragraph discuss how the positive and negative samples are formed.}
%

% \xy{Some example:}

% \noindent{\textbf{Data Sampling:}} 
% Consider one modality, i.e. RGB, in the target domain, to collect the positive sample pairs, we apply the source domain trained classifier to pseudo label the target domain video frames. Then, we select those sample frames with pseudo labeling score larger than certain threshold $T$ (e.g., $T=0.8$ in our setting) to form the positive pairs. Those smaller than the threshold samples together with the samples whose score is larger the threshold, are paired as the negative pairs. (there are actually other pairing, cross source and target domain, etc.)

\vspace{-4mm}
\paragraph{Similarity Between Samples.}
To measure the sample similarity in our contrastive objective, we adopt $\phi^{st}$ to calculate the similarity between cross-domain features:
% Here, $\phi^{st}$ measures the similarity between cross-domain features:
%
%\BB{Why the cross-modality loss is split into two terms according to domain, but here the cross-domain loss is not split into two terms according to modality?}
\begin{equation}
    \label{eq:sim_st}
    \phi^{st}(F_{t_i}^k, F_{s_i}^k)_{k \: \in \{a, m\}} = \text{exp}( F_{t_i}^k{^\top} F_{s_i}^k / \tau ),
\end{equation}
where the modality $k$ can be appearance or motion in our work.
Note that, for cross-domain feature regularization, in order to align features, we do not use an additional projection head $h(\cdot)$ like Section \ref{sec:mo} or Figure \ref{fig:sampling_mo} (explained as the follows).
%

%\vspace{-4mm}
\paragraph{Loss Function.}
The corresponding loss function with respect to the sampling strategy (Figure \ref{fig:sampling_do}) and the similarity measurement $\phi^{st}$ for the RGB modality is defined as:
% In Figure \ref{fig:sampling_do}, we illustrate this process for the source and target features considering the RGB modality as one example, and the corresponding loss function is defined as:
%
\begin{equation}
    \label{eq:do_a}
    \mathcal{L}_{do}^a = - \text{log} \frac{\sum\limits_{t_i \in \hat{V_t}} \phi^{st}(F_{t_i}^a, F_{s_i+}^a) }{ \sum\limits_{t_i \in \hat{V_t}} \phi^{st}(F_{t_i}^a, F_{s_i+}^a) + \phi^{st}(F_{t_i}^a, F_{s_i-}^a)},
\end{equation}
%
% \begin{equation}
%     \label{eq:do}
%     \mathcal{L}_{do} = - \text{log} \frac{\sum\limits_{k \: \in \{a, m\}} \phi^{st}(F_{t_i}^k, F_{s_i+}^k) }{ \sum\limits_{k \: \in \{a, m\}} \phi^{st}(F_{t_i}^k, F_{s_i+}^k) + \phi^{st}(F_{t_i}^k, F_{s_i-}^k)},
% \end{equation}
%\BB{If we avoid using k.l in Eq.(1), Eq.(4) can be simpler, by using $k\in\{a,m\}$ and $(F_{t_i}^k, F_{s_i+}^k)$ etc.}
where $t_i \in \hat{V_t}$ is the set of target videos with a more confident pseudo labels $\hat{Y_t}$. $F_{s_i+}^a$ are the positive samples of source videos $s_i \in V_s$ with the same class label as $\hat{Y_t}$, while $F_{s_i-}^a$ are the negative samples with different class labels.
% with respect to the source positive/negative video set $s_i+$ and $s_i-$.
Similarly, when the modality is flow, the loss function is:
\begin{equation}
    \label{eq:do_m}
    \mathcal{L}_{do}^m = - \text{log} \frac{\sum\limits_{t_i \in \hat{V_t}} \phi^{st}(F_{t_i}^m, F_{s_i+}^m) }{ \sum\limits_{t_i \in \hat{V_t}} \phi^{st}(F_{t_i}^m, F_{s_i+}^m) + \phi^{st}(F_{t_i}^m, F_{s_i-}^m)}.
\end{equation}
We also note that the choice without using the projection head $h(\cdot)$ is reasonable as our objectives in \eqref{eq:do_a} and \eqref{eq:do_m} essentially try to make features with the same action labels closer to each other, which is consistent with the final goal for performing action recognition.

% \vspace{-2mm}
% \paragraph{Sampling Strategy.} \YH{TODO} Given a target RGB/flow feature, we give a pseudo label if the confidence is higher than a threshold $T$. We treat all source RGB/flow features with the same label as the pseudo label as positive features and treat all other features with different labels as 
% negative features.

\vspace{-4mm}
\paragraph{Connections to Pseudo-label Self-training.}
Using pseudo labels on the target sample to self-train the model is one commonly used approach in domain adaptation \cite{Zou_iccv19,Zou_ECCV_2018,Lian_ICCV19}.
In the proposed cross-domain contrastive learning, we also adopt pseudo labels to form positive samples. However, these two methods are distinct, in terms of the way that reshapes the feature space.

Given the target video $V_t$, one can produce a pseudo-label $\hat{Y_t}$ and use it for training the action recognition network with the standard cross-entropy loss.
Therefore, such supervision is a strong signal that forces the feature $F_t$ to map into the space of action label $\hat{Y_t}$, which is sensitive to noisy labels such as pseudo-labels.
On the contrary, using contrastive learning with pseudo-labels is similar in spirit to the soft nearest-neighbor loss \cite{pmlr-v2-salakhutdinov07a,Wu_eccv18}, which encourages soft feature alignments, rather than enforcing the hard final classification, hence more robust to potential erroneous pseudo-labels.
%
% \BB{The reason of higher robustness of contrastive learning should be more clearly and  explicitly argued, rather than just refering to soft NN loss. Readers who do not know this loss may not understand, or need to read [39,57]. For example, can we say "Similar in spirit to the soft nearest-neighbor loss \cite{pmlr-v2-salakhutdinov07a,Wu_eccv18}, contrastive learning with pseudo labels encourages soft feature alignments, rather than enforcing the hard final classification, hence more robust to potential wrong pseudo-labels."?}
Similar observations are also presented in the recent work, which shows that the supervised contrastive loss \cite{Khosla_nips_20} is more robust than the cross-entropy loss in image classification.
In our case, we utilize such property and show that cross-domain contrastive learning can be achieved by using pseudo-labels in video domain adaptation, and is more robust than pseudo-label self-training. More empirical comparisons will follow in the experiments.
% \xy{maybe could be more concise. The central meaning is: contrastive learning with pseudo label is more following the soft nearest neighbor spirit and could potentially be more robust than pseudo-label self-training. The empirical evaluation following will support the statement.}

\subsection{A Contrastive Learning Framework}
In previous sections, we have introduced how we incorporate cross-modal and cross-domain contrastive loss functions to regularize features extracted from RGB/flow branches across the source and target domains. Next, we introduce the entire objective.

\vspace{-4mm}
\paragraph{Overall Objective.}
Overall, we include loss functions in Section \ref{sec:mo} and \ref{sec:cross_domain} without using any supervisions, and a standard supervised cross-entropy loss $\mathcal{L}_{src}$ on source videos $V_s$ with action labels $Y_s$. To obtain the final output from the two-stream network, we average the outputs from individual classifiers of RGB and flow branches (see Figure \ref{fig:overview}).
% \vspace{-4mm}
\begin{equation}
    \label{eq:obj}
    \begin{split}
    \mathcal{L}_{all} & = \mathcal{L}_{src}(V_s, Y_s) + \\
    & \lambda (\mathcal{L}_{mo}^s(V_s) + \mathcal{L}_{mo}^t(V_t) + \mathcal{L}_{do}(V_s, V_t,\hat{Y}_t) ),
    \end{split}
\end{equation}
%
% \begin{equation}
%     \label{eq:obj}
%     \begin{split}
%     \mathcal{L}_{all} & = \mathcal{L}_{src}(V_s, Y_s) + \lambda (\mathcal{L}_{mo}^s(V_s) + \mathcal{L}_{mo}^t(V_t)) + \\
%     & \lambda (\mathcal{L}_{do}^a(V_s, V_t,\hat{Y}_t) + \mathcal{L}_{do}^m(V_s, V_t,\hat{Y}_t)),
%     \end{split}
% \end{equation}
where $\lambda$ is the weight to balance the terms. Here, we treat loss functions in \eqref{eq:do_a} and \eqref{eq:do_m} as one term: $\mathcal{L}_{do} = \mathcal{L}_{do}^a + \mathcal{L}_{do}^m$, as they are with the same form but using different modalities.
%
% $\lambda$ is the weight to balance the cross-modality and cross-domain losses (\ie, $\lambda = 0.1$ in this paper).
%
Since all the loss terms are with a similar form, it does not require heavy tuning on each of them, so that we use the same $\lambda$ for cross-modal and cross-domain losses (\ie, $\lambda = 1.25$ in this paper).

% Note that $\mathcal{L}_{mo}^s$ and $\mathcal{L}_{mo}^t$ are in the same loss form, while $\mathcal{L}_{co}$ takes videos in two domains at the same time and is the same form for either the RGB or the flow feature.

% \paragraph{Overall Objective.}
\vspace{-4mm}
\paragraph{Leveraging Memory Banks.}
% \subsection{Leveraging Memory Banks}
To compute the cross-modal and cross-domain loss functions, we need to compute all feature representations summed from video sets $V_s$ and $V_t$. However, it is impossible to obtain all the features at every training iteration. Therefore, we store the features in the memory banks following \cite{wu2018unsupervised}, i.e., an individual memory bank for a domain and a modality, totally with four combinations $M_s^a, M_s^m, M_t^a,$ and $M_t^m$. Given the features in a batch (\ie, $F_{s_i}^a, F_{s_i}^m, F_{t_j}^a, $ and $F_{t_j}^m$), we pull out features from the memory banks for positive and negative features (\eg, $F_{s_i+}^a$ or $F_{s_i-}^a$ in \eqref{eq:do_a} is replaced by $M_{s_i+}^a$ and  $M_{s_i-}^a$). Then, the memory bank features are updated with the features in the batch at the end of each iteration. We use a momentum update $\delta=0.5$ following \cite{wu2018unsupervised}:
\begin{equation}
    M_{s_i}^a = \delta M_{s_i}^a + (1-\delta)F_{s_i}^a.
\end{equation}
% \mc{MC: Did you do some validation for the momentum value? This seems quite different from MoCo which sets it very close to 1.}
The other memory banks, $M_s^m, M_t^a,$ and $M_t^m$, are also updated in the same way. Using the momentum update also encourages smoothness in training dynamics~\cite{wu2018unsupervised}.
In our case, during the training process, we randomly sample consecutive frames in a video clip. Therefore, by using the memory banks, our model can also encourage the temporal smoothness within each clip in feature learning.
% \paragraph{Implementation Details.}
% - backbone architecture, contrastive learning, pseudo-labels

\section{Experimental Results}

% - \dennis{TODO: (running, almost done)} UCF to HMDB, final w/ adv (at least have this one), final w/o adv, cross-modality, cross-domain

% - \dennis{TODO: (running)} pseudo-label retraining for Kitchen

% - \dennis{TODO: (done)} update new results for Kitchen in Table 2, but if it does not seem good, we can still use the current results

% - \dennis{TODO: (running)} add results in Table 3

% - \dennis{TODO: (done)} More analysis, like t-SNE plots on four feature spaces, compare with pseudo-labeling self-training

% - \dennis{TODO: (high priority)} Pick video clips to show successful examples for cross-modality and cross-domain effects, show feature similarities to source videos

% - \dennis{TODO: (lower priority)} top-K accuracy \\

%In this section, we first introduce the dataset in our experiments. Then, we show performance comparisons on numerous domain adaptation benchmark scenarios for action recognition. Furthermore, we provide comprehensive ablation studies to validate the effectiveness of our cross-domain and cross-modal feature regularization. Please refer to the supplementary material for more results and analysis.

In this section, we show performance comparisons on numerous domain adaptation benchmark scenarios for action recognition, followed by comprehensive ablation studies to validate the effectiveness of our cross-domain and cross-modal feature regularization. Please refer to the supplementary material for more results and analysis.

% Finally, we present a few analysis and sensitivity studies to show the robustness of our proposed framework.

\subsection{Datasets and Experimental Setting}

We use the three standard benchmark datasets for video domain adaptation, UCF \cite{ucf}, HMDB \cite{hmdb}, and EPIC-Kitchens \cite{epic-kitchen}. We then show that our method is a general framework to work for different types of action recognition settings: UCF $\leftrightarrow$ HMDB for human activity recognition, as well as EPIC-Kitchens for fine-grained action recognition in egocentric videos.
 
\vspace{-4mm}
\paragraph{UCF $\leftrightarrow$ HMDB.}
Chen et al. \cite{Chen_da_iccv19} release the UCF $\leftrightarrow$ HMDB dataset for studying video domain adaptation. This dataset has $3209$ videos with 12 action classes.
All the videos come from the original UCF \cite{ucf} and HMDB \cite{hmdb} datasets, which sample the overlapping 12 classes out of 101/51 classes from UCF/HMDB,
respectively. There are two settings of interest, UCF $\rightarrow$ HMDB and HMDB $\rightarrow$ UCF. We show the performance of our method in both settings following the official split provided by the authors \cite{Chen_da_iccv19}.

\vspace{-4mm}
\paragraph{EPIC-Kitchens.}
This dataset contains fine-grained action classes with videos recorded in different kitchens from the egocentric view. We follow the train/test split used in \cite{Munro_cvpr20} for the domain adaptation task. There are 8 action categories in the three largest kitchens, \ie,  D1, D2, and D3, and we use all the pairs of them as source/target domains. Note that, compared to UCF $\leftrightarrow$ HMDB, EPIC-Kitchens is more challenging as it has more fined-grained classes (\eg, ``Take'', ``Put'', ``Open'', ``Close'') and imbalanced class distributions. We report the top-1 accuracy on the test set averaged over the last 9 epochs following \cite{Munro_cvpr20}.

% \textbf{UCF-HMDB} contains 3,209 video clips with 12 action classes from UCF and HMDB. For video domain adaptation, 12 subclasses are sampled from out of the original dataset. This benchmark contains less domain gap than the EPIC-Kitchens dataset.
% We use the same split by []. 

% \textbf{EPIC-Kitchens} contains domain adaptation splits with 8 classes of different domains recorded from different kitchens. Compared to UCF-HMDB, EPIC-Kitchens is more fined-grained classification (\eg, `Take', `Put', and etc.) and has imbalanced class distribution. We report top-1 target accuracy on the test set of the target domain.
\begin{table}[!t]
	\caption{
		Performance comparisons on UCF $\leftrightarrow$ HMDB.
	}
% 	\vspace{1mm}
	\label{table:ucf}
	\small
	\centering
	\renewcommand{\arraystretch}{1.0}
	\setlength{\tabcolsep}{3pt}
	\resizebox{0.4\textwidth}{!}
	{\begin{tabular}{lccc}
		\toprule
		
		Setting & Two-stream & UCF $\rightarrow$ HMDB & HMDB $\rightarrow$ UCF \\

        \midrule
		
		Source-only \cite{Chen_da_iccv19} &  & 80.6 & 88.8 \\
		
		TA$^3$N \cite{Chen_da_iccv19} & & 81.4 & 90.5 \\
		
		Supervised-target \cite{Chen_da_iccv19} & & 93.1 & 97.0 \\
		
		\midrule
		
		TCoN \cite{Pan_aaai20} & & \textbf{87.2} & 89.1 \\
		
		\midrule
		
		Source-only \cite{Choi_eccv20} & & 80.3 & 88.8 \\
		
		SAVA \cite{Choi_eccv20} & & 82.2 & \textbf{91.2} \\
		
		Supervised-target \cite{Choi_eccv20} & & 95.0 & 96.8 \\
		
		\midrule
		
% 		\midrule
		
		Source-only  & $\surd$ & 82.8 & 90.7 \\
		
		MM-SADA \cite{Munro_cvpr20} & $\surd$ & 84.2 & 91.1 \\
		
		Ours (cross-modal) & $\surd$ & \textbf{84.7} & 92.5 \\
		
		Ours (cross-domain) & $\surd$ & 83.6 & 91.1 \\
		
		Ours (final) & $\surd$ & \textbf{84.7} & \textbf{92.8} \\
		
% 		Ours (final w/ adv) & \dennis{TODO} & \\
		
		Supervised-target  & $\surd$ & 98.8 & 95.0\\
		
		\bottomrule
	\end{tabular}}
	\vspace{-4mm}
\end{table}

\begin{table*}[!t]
    \begin{minipage}[t]{.7\linewidth}
	\caption{
		Performance comparisons on EPIC-Kitchens.
	}
% 	\vspace{1mm}
	\label{table:kitchen}
	\small
    % \scriptsize
	\centering
	\renewcommand{\arraystretch}{1.0}
	\setlength{\tabcolsep}{3pt}
	\resizebox{0.95\textwidth}{!}
	{\begin{tabular}{lcccccccc}
		\toprule
		
		Setting & D2 $\rightarrow$ D1 & D3 $\rightarrow$ D1 & D1 $\rightarrow$ D2 & D3 $\rightarrow$ D2 & D1 $\rightarrow$ D3 & D2 $\rightarrow$ D3 & Mean & Gain\\

        \midrule
		
		Source-only & 42.5 & 44.3 & 42.0 & 56.3 & 41.2 & 46.5 & 45.5 \\
		
		\midrule
		
		AdaBN \cite{adabn}  & 44.6  & 47.8  & 47.0  & 54.7  & 40.3  & 48.8  & 47.2 & +1.7 \\
		
		MMD \cite{long2015learning}  & 43.1  & 48.3  & 46.6  & 55.2  & 39.2  & 48.5  & 46.8 & +1.3 \\
		
		MCD \cite{Saito_CVPR_2018}  & 42.1  & 47.9  & 46.5  & 52.7  & 43.5  & 51.0  & 47.3 & +1.8 \\
		
% 		MM-SADA \cite{Munro_cvpr20} & 41.8 & 49.7 & 47.7 & 57.4 & 40.3 & 50.6 & 47.9 & +2.4 \\
		
% 		MM-SADA \cite{Munro_cvpr20} & 48.2 & 50.9 & 49.5 & 56.1 & 44.1 & 52.7 & 50.3 & +4.8 \\
		
		MM-SADA \cite{Munro_cvpr20} & 47.4 & 48.6 & 50.8 & \textbf{56.9} & 42.5 & \textbf{53.3} & 49.9 & +4.4 \\
		
% 		\midrule
		
		Ours (modality) & 44.3 & 50.2 & 49.5 & 56.6 & 43.0 & 48.8 & 48.7 & +3.2\\
		
		Ours (domain) & 47.4 & \textbf{52.8} & \textbf{52.4} & 56.1 & 41.7 & 49.9 & 50.1 & +4.6 \\
		
		Ours (final) & \textbf{49.5} & 51.5 & 50.3 & 56.3 & \textbf{46.3} & 52.0 & \textbf{51.0} & \textbf{+5.5} \\

		\midrule
		Supervised-target & 62.8 & 62.8 & 71.7 & 71.7 & 74.0 & 74.0 & 69.5 \\
		
		\bottomrule
	\end{tabular}}
% 	\vspace{-4mm}
	\end{minipage}
	\hfill
	\begin{minipage}[t]{.3\linewidth}
	\caption{
		Ablation study on EPIC-Kitchens.
	}
% 	\vspace{1mm}
	\label{table:kitchen_ablation}
	\small
% 	\scriptsize
	\centering
	\renewcommand{\arraystretch}{1.1}
	\setlength{\tabcolsep}{3pt}
	\resizebox{\textwidth}{!}
	{\begin{tabular}{lcccc}
		\toprule
		
		Setting & Modality & Domain & Mean & Gain \\

		\midrule
		
		Source-only & & & 45.5 &  \\
		
		\midrule
		
		MM-SADA \cite{Munro_cvpr20} & $\surd$ & & 47.9 & +2.4 \\
		
		Ours & $\surd$ & & \textbf{48.7} & \textbf{+3.2} \\
		
		\midrule
		
		MM-SADA \cite{Munro_cvpr20} & & $\surd$ & 49.4 & +3.9 \\
		
		Pseudo-labeling & & $\surd$ & 49.0 & +3.5 \\
		
		Ours & & $\surd$ & \textbf{50.1} & \textbf{+4.6} \\
		
		\midrule
		
		MM-SADA \cite{Munro_cvpr20} & $\surd$ & $\surd$ & 49.9 & +4.4 \\
		
		Ours & $\surd$ & $\surd$ & \textbf{51.0} & \textbf{+5.5} \\
		
% 		\midrule
		
% 		Supervised-target  & 69.5 & \\
		
		\bottomrule
	\end{tabular}}
	\end{minipage}
	
	\vspace{-4mm}
\end{table*}

% \begin{table*}[!t]
% 	\caption{
% 		Performance comparisons on EPIC-Kitchens.
% 	}
% % 	\vspace{1mm}
% 	\label{table:kitchen}
% 	\small
% 	\centering
% 	\renewcommand{\arraystretch}{1.1}
% 	\setlength{\tabcolsep}{5pt}
% 	\resizebox{0.8\textwidth}{!}{\begin{tabular}{lcccccccc}
% 		\toprule
		
% 		Setting & D2 $\rightarrow$ D1 & D3 $\rightarrow$ D1 & D1 $\rightarrow$ D2 & D3 $\rightarrow$ D2 & D1 $\rightarrow$ D3 & D2 $\rightarrow$ D3 & Mean & Gain\\

%         \midrule
		
% 		Source-only & 42.5 & 44.3 & 42.0 & 56.3 & 41.2 & 46.5 & 45.5 \\
		
% 		\midrule
		
% 		AdaBN \cite{adabn}  & 44.6  & 47.8  & 47.0  & 54.7  & 40.3  & 48.8  & 47.2 & +1.7 \\
		
% 		MMD \cite{long2015learning}  & 43.1  & 48.3  & 46.6  & 55.2  & 39.2  & 48.5  & 46.8 & +1.3 \\
		
% 		MCD \cite{Saito_CVPR_2018}  & 42.1  & 47.9  & 46.5  & 52.7  & 43.5  & 51.0  & 47.3 & +1.8 \\
		
% % 		MM-SADA \cite{Munro_cvpr20} & 41.8 & 49.7 & 47.7 & 57.4 & 40.3 & 50.6 & 47.9 & +2.4 \\
		
% % 		MM-SADA \cite{Munro_cvpr20} & 48.2 & 50.9 & 49.5 & 56.1 & 44.1 & 52.7 & 50.3 & +4.8 \\
		
% 		MM-SADA \cite{Munro_cvpr20} & 47.4 & 48.6 & 50.8 & \textbf{56.9} & 42.5 & \textbf{53.3} & 49.9 & +4.4 \\
		
% % 		\midrule
		
% 		Ours (cross-modality) & 44.3 & 50.2 & 49.5 & 56.6 & 43.0 & 48.8 & 48.7 & +3.2\\
		
% 		Ours (cross-domain) & 47.4 & \textbf{52.8} & \textbf{52.4} & 56.1 & 41.7 & 49.9 & 50.1 & +4.6 \\
		
% 		Ours & \textbf{49.5} & 51.5 & 50.3 & 56.3 & \textbf{46.3} & 52.0 & \textbf{51.0} & \textbf{+5.5} \\

% 		\midrule
% 		Supervised-target & 62.8 & 62.8 & 71.7 & 71.7 & 74.0 & 74.0 & 69.5 \\
		
% 		\bottomrule
% 	\end{tabular}}
% 	\vspace{-3mm}
% \end{table*}

\vspace{-4mm}
\paragraph{Implementation Details.}
Our entire framework is implemented in PyTorch using 2 TITANXP GPUs. We use an I3D two-stream network \cite{Carreira_cvpr17} composed of an RGB stream and a flow stream, where the network is pre-trained on Kinetics following \cite{Choi_eccv20}. During training, we use the same setting as \cite{Choi_eccv20, Munro_cvpr20} to randomly sample 16 consecutive frames out of a video clip. Then each RGB and flow stream takes these 16 frames with a size of 224$\times$224. Each stream is followed by a fully-connected layer to compute individual output logits. Then the logits from each stream are averaged to predict the final class scores. To optimize the entire network, we use the SGD optimizer with a learning rate of 0.01. We set the temperature $\tau=0.1$ and $\delta=0.5$ for all experiments following \cite{wu2018unsupervised}. For UCF $\leftrightarrow$ HMDB, we follow the setting as \cite{Choi_eccv20} for batch size, total training epochs, learning rate, etc. For EPIC-Kitchens, we implement it upon the official code of~\cite{Munro_cvpr20} but set the batch size as 32 to fit the memory of 2 GPUs and train the model for 6K iterations. The learning rate decreases by a factor of 10 every 3K iterations.
% \YH{How do we set the batch size as 32 for EPIC-Kitchens? We may update some descriptions as you are now using the implementation of \cite{Munro_cvpr20} and we need to state the difference in GPU numbers.}
% During testing, we use the sample the center clip with 16 frames. 

\subsection{Results on UCF $\leftrightarrow$ HMDB}
We show experimental results for UCF $\rightarrow$ HMDB and HMDB $\rightarrow$ UCF in Table \ref{table:ucf}, comparing with state-of-the-art methods --- TA$^3$N \cite{Chen_da_iccv19}, TCoN~\cite{Pan_aaai20}, SAVA \cite{Choi_eccv20}, and MM-SADA \cite{Munro_cvpr20}.

\vspace{-4mm}
\paragraph{Comparisons with State-of-the-art Methods.}
In each group of Table \ref{table:ucf}, in addition to the result for each method, we show the ``Source-only'' model that only trains on videos in the source domain, and the ``Supervised-target'' model that trains on target videos with ground truths, which serves as an upper bound. We also implement~\cite{Munro_cvpr20} on UCF-HMDB in the same setup for fair comparisons.
Different from the TA$^3$N, TCoN, and SAVA that only exploit the single modality via adversarial feature alignment and self-learning schemes, our method leverages both the RGB and flow modalities in a domain adaptation framework, which achieves the state-of-the-art performance.
We also notice that our source-only model performs slightly better than the other source-only baselines, due to the usage of the flow stream. Despite that the domain gap is reduced by leveraging the flow modality, our approach still obtains comparable performance gains with respect to the source-only model and performs better than MM-SADA \cite{Munro_cvpr20} that adopts the same two-stream model.
% \YH{We may need to note that we implement the MM-SADA model using our implementation for fair comparisons.}
For example, on UCF $\rightarrow$ HMDB, the gain for TA$^3$N and SAVA is 0.8\% and 1.9\%, respectively, while our gain is the same as SAVA and is much higher than TA$^3$N.

\vspace{-4mm}
\paragraph{Ablation Study.}
In the fourth group of Table \ref{table:ucf}, we show model variants to validate the usefulness of individual components in our contrastive learning framework, \ie, cross-modal and cross-domain feature regularization.
From the results, the two modules consistently improve the performance over the source-only baseline. By combining both, it provides the highest accuracy.
Here, interestingly, we find that the cross-domain module is less helpful than the cross-modal one. One reason is that on UCF $\leftrightarrow$ HMDB, these two domains already share a high similarity, which reduces the impact of using the cross-domain loss. However, this also shows the importance of incorporating the proposed two modules, in which the cross-modal loss can still provide effective regularization, even when the domain gap is smaller. In the next section, we will show a different scenario, where both modules are important.

\begin{figure*}[!t]
	\centering
	\vspace{-0.3cm}
	\includegraphics[width=0.9\linewidth]{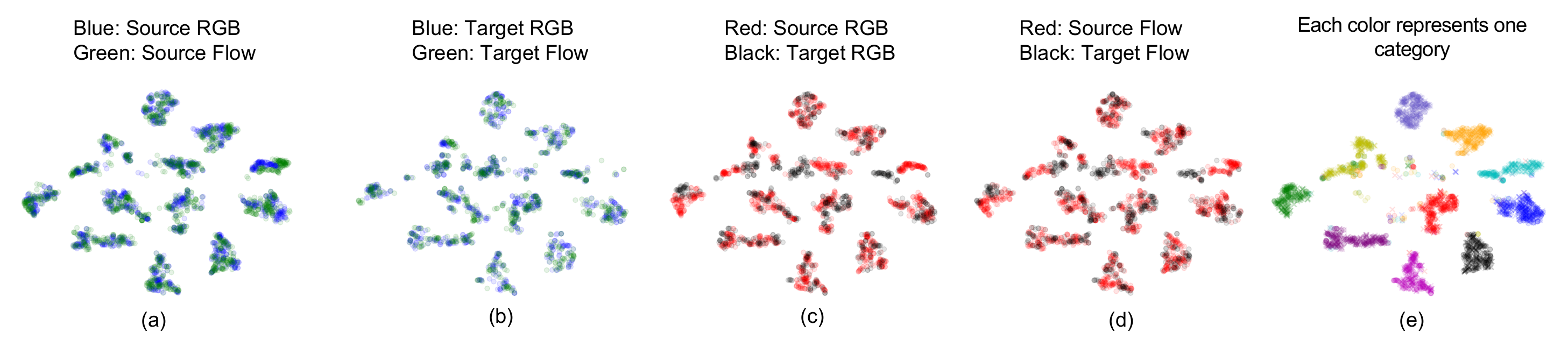}
	\vspace{-1mm}
	\caption{
	t-SNE visualization on cross-modal and cross-domain features after the projection head $h(\cdot)$ on UCF $\rightarrow$ HMDB, \ie $h(F^a_s)$, $h(F^m_s)$, $h(F^a_t)$, $h(F^m_t)$. In (a)(b), we show the visualization for individual domains, where each domain contains the multi-modality features. In (c)(d), we visualize features for each modality, and each plot uses the features from two domains. (e) includes all the features from two domains and two modalities, where each color represents one action class.
	}
	\label{fig:tsne}
	\vspace{-3mm}
\end{figure*}

\subsection{Results on EPIC-Kitchens}

We present results on the EPIC-Kitchens benchmark for domain adaptation \cite{Munro_cvpr20}, including comparisons with state-of-the-art methods, ablation study, and more analysis.

\vspace{-4mm}
\paragraph{Comparisons with State-of-the-art Methods.}
In Table \ref{table:kitchen}, we present several domain adaptation methods, including distribution alignment via adversarial learning \cite{long2015learning}, maximum classifier discrepancy \cite{Saito_CVPR_2018}, and adaptive batch normalization \cite{adabn}, and a recently proposed method that uses a self-learning objective \cite{Munro_cvpr20}.
We note that these results are reported from \cite{Munro_cvpr20} using the same two-stream feature extractor as ours, and they share the same ``Source-only'' model and ``Supervised-target'' upper bound in Table \ref{table:kitchen}.
For fair comparisons, we reproduce results of MM-SADA~\cite{Munro_cvpr20} using their official implementation and the same computing resources as ours, and show that our final model performs better than MM-SADA by 1.1\% on average. The results show the advantages of our contrastive learning framework. More detailed analysis is provided as follows.

% We also note that, in most kitchen-pair settings, our method performs favorably against \cite{Munro_cvpr20}, except for the ``D3 $\rightarrow$ D2'' setting. In this setting, our source-only model fails to produce better performance, although our performance gain (\ie, 2.5\%) is larger than that in \cite{Munro_cvpr20} as 1.1\%.

%
% Moreover, compared to the upper bound results, our method achieves the average performance closer to the upper bound, which is one indicator to demonstrate a better adaptation ability.

\vspace{-4mm}
\paragraph{Ablation Study.}
In Table \ref{table:kitchen_ablation}, we ablate the two components for our cross-modal and cross-domain loss functions with other approaches that consider the similar aspect.
For fair comparisons, we use the same two-stream backbone, implementation, and computing resources for generating all the results.
Considering only modality or domain, our method consistently performs better than MM-SADA~\cite{Munro_cvpr20}, in which it uses a self-learning module to predict whether the RGB/flow modality comes from the same video clip and a typical adversarial learning scheme to align cross-domain features.
Combining these two factors, our method improves the ``Source-only'' model the most, which shows the effectiveness of the proposed unified framework using contrastive learning.
In addition, it is worth mentioning that our cross-domain loss performs better than pseudo-label self-training by $1.1$\%, which validates the discussion in Section \ref{sec:cross_domain} on the difference of leveraging pseudo-labels.

\vspace{-4mm}
\paragraph{Sampling Strategy.}
In Table \ref{table:kitchen_sampling}, we present the ablation for sampling strategy in the cross-modal loss (see Section \ref{sec:mo}), where we do not assume that the RGB and flow modality have the same window of frames, which handles the large intra-variation within a video clip.
When applying this strategy in MM-SADA~\cite{Munro_cvpr20}, acting as a way for data augmentation, the performance gain is smaller than ours (\ie, 0.5\% vs 1.1\%).
This validates our sampling strategy with the proposed contrastive learning objective that enriches feature regularization under the domain adaptation setting.

\begin{figure}[!t]
	\centering
	\includegraphics[width=\linewidth]{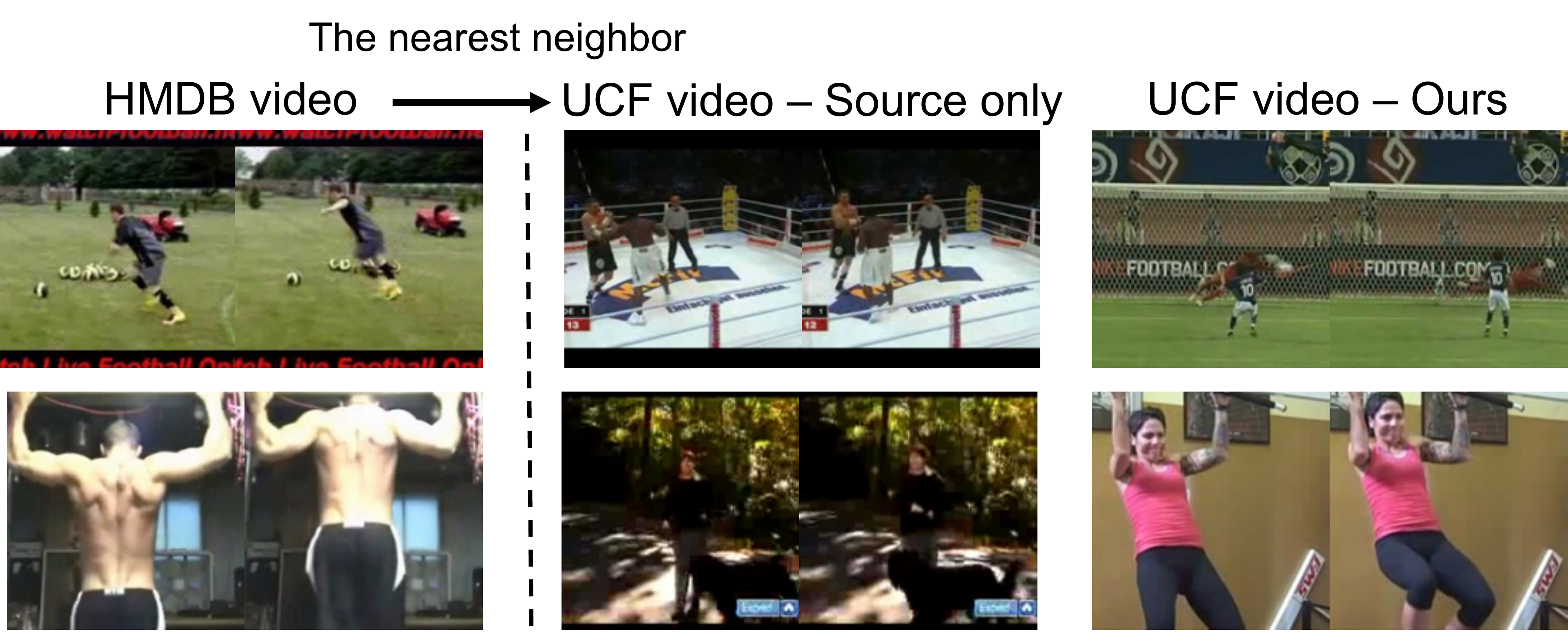}
% 	\vspace{-4mm}
	\caption{Cross-domain retrievals using the RGB feature. Given the target feature $F_t^a$, we retrieve the closest neighbor $F_s^a$ in the source domain. Our model correctly aligns videos of the same class under view-angle (1st row) and background (2nd row) differences.
% 	, while the source-only model is biased to the background context.
	}
	\label{fig:retreival}
	\vspace{-4mm}
\end{figure}

\subsection{More Results and Analysis}

%We present more analysis including the feature visualizations using t-SNE \cite{tsne}, and some example results for action recognition to understand where our model makes the correct prediction.

We present more analysis including the feature visualizations using t-SNE \cite{tsne}, and example results for cross-domain retrieval to understand our model predictions.

\vspace{-4mm}
\paragraph{t-SNE Visualizations.}
% In this paper, we propose to regularize the cross-modality via a contrastive loss function through a projection head $h(\cdot)$ via \eqref{eq:mo}.
%
In this paper, we use a projection head $h(\cdot)$ to project RGB/flow features to an embedding space for cross-modal loss in our framework.
Therefore, it is of great interest to understand how the features behave in this embedding space. To this end, we samples features from both domains across two modalities, and perform t-SNE visualizations on UCF $\rightarrow$ HMDB in Figure \ref{fig:tsne}.

Here, although our cross-modal loss is calculated in each domain, the projection head is shared across the domains.
Therefore, we provide different combinations of the feature spaces to visualize how the four feature spaces look like, \ie, $h(F^a_s)$, $h(F^m_s)$, $h(F^a_t)$, $h(F^m_t)$.
First, in Figure \ref{fig:tsne}(a)(b), it is not surprising to observe that in each domain, features from different modalities are aligned together (\eg, Source RGB/Flow in Figure \ref{fig:tsne}(a)), as it is exactly what the cross-modal objective in \eqref{eq:mo} optimizes for.

\begin{table}[!t]
	\caption{
		Ablation study on the sampling strategy.
	}
% 	\vspace{1mm}
	\label{table:kitchen_sampling}
	\small
	\centering
	\renewcommand{\arraystretch}{1.0}
	\setlength{\tabcolsep}{3pt}
	\resizebox{0.35\textwidth}{!}{\begin{tabular}{lccc}
		\toprule
		
		Setting & Sampling & Mean & Gain via sampling \\

% 		\midrule
		
% 		Source-only & 45.5 &  \\
		
		\midrule
		
		\multirow{2}{*}{MM-SADA \cite{Munro_cvpr20}} & & 49.4 & \\
		
		& $\surd$ & 49.9 & +0.5 \\
		
		\midrule
		
		\multirow{2}{*}{Ours} & & 49.9 &  \\
		
		& $\surd$ & 51.0 & +1.1 \\
		
% 		\midrule
		
% 		Supervised-target  & 69.5 & \\
		
		\bottomrule
	\end{tabular}}
	\vspace{-5mm}
\end{table}

More interestingly, if we consider each modality at a time in Figure \ref{fig:tsne}(c)(d), \eg, Source RGB and Target RGB, their features are also aligned well, even we do not explicitly have an objective to align them in the embedding space via $h(\cdot)$.
This shows the merit of our framework that enables feature regularization and interplay across four feature spaces.
% This shows the merit of introducing the cross-domain loss
% % \BB{I thought the flow does not appear here, so nothing to do with cross-modality? If source RGB and target RGB are aligned, that's due to cross-domain loss?}
% , in which features are aligned in an embedding space between the source and target features, and such behavior further provides a way to regularize features in the backbone network for learning domain adaptive representations.
% 
Also, we present a visualization on the distribution for each class, including all the source and target features across two modalities in Figure \ref{fig:tsne}(e). This illustrates that features from the same category are aligned well.
% , which provides useful supervisory signals to learn domain-invariant features.

\vspace{-5mm}
\paragraph{Cross-domain Retrievals.} In Figure~\ref{fig:retreival}, we show the cross-domain video retrievals using the RGB feature. Based on the target feature in HMDB, we show the nearest neighbor one from UCF. We show that our method can correctly retrieve the videos of the same class, either having the same context background but from a different view-angle or acting in a similar movement but with a different background.

% - Action recognition datasets: UCF $\rightarrow$ HMDB, HMDB $\rightarrow$ UCF (already saturated), Kitchen, Kinetics $\rightarrow$ NEC Drone

% - Baselines (I3D): source-only, supervised-target (upper bound), SOTA

% - Ablation: + cross-modality CL, + cross-domain CL, ours, ours without using the projection head for CL

% \begin{table}[!h]
% 	\caption{
% 		Ablation study.
% 	}
% % 	\vspace{1mm}
% 	\label{table:baseline}
% 	\small
% 	\centering
% 	\renewcommand{\arraystretch}{1.1}
% 	\setlength{\tabcolsep}{3pt}
% 	\begin{tabular}{lcc}
% 		\toprule
		
% 		Setting & UCF $\rightarrow$ HMDB & Kitchen \\

%         \midrule
		
% 		Source-only & 80.3 &  44.9 \\
		
% 		Ours w/ cross-modality CL  &  87.5  & 47.7 \\
		
% 		Ours w/ cross-domain CL  &  & \\
		
% % 		Ours w/o projection head &  & \\
		
% 		Ours (final) &  & \\
		
% 		\midrule
		
% 		Supervised-target & 95.0 &  \\
		
% 		\bottomrule
% 	\end{tabular}
% % 	\vspace{-3mm}
% \end{table}

% Analysis: in cross-domain CL, there are hyperparameters, e.g.., thresholds for pseudo-labels, top $K$ samples via the feature similarity

% \vspace{-2mm}
\section{Conclusions}
% \vspace{-1mm}
We investigate the video domain adaptation task with our cross-modal contrastive learning framework.
% Previous video domain adaptation methods try to find transferable temporal or spatial region to align the source and target domains. 
To this end, we leverage the multi-modal, RGB and flow information, and exploit their relationships.
In order to handle feature spaces across modalities and domains, we propose two objectives to regularize such feature spaces, namely cross-modal and cross-domain contrastive losses, that learn better feature representations for domain adaptive action recognition.
%
% which learn better knowledge of RGB and flow streams with cross-modality contrastive learning and align the source and target features with cross-domain contrastive learning.
Moreover, our framework is modular, so it can be applicable to other domain adaptive multi-modal applications, which will be considered as the future work.

{\small
\bibliographystyle{ieee_fullname}
\bibliography{egbib}
}

\appendix
\section*{Appendix}
\section{Implementation Details}
We use 2 TITANXP GPUs in our implementation. We also reproduce the results of MM-SADA~\cite{Munro_cvpr20}\footnote{https://github.com/jonmun/MM-SADA-code} using their released code with the same 2-GPU setup and the same batch size as our method.

\section{Ablation study}

% We show the ablation study of our method in Tables~\ref{table:ablation_hyper_params} and~\ref{table:ablation_projection_head}.
In Table~\ref{table:ablation_hyper_params}, we show the sensitivity analysis on the $\lambda$ (Eq. (7) in the main paper) and the confidence threshold $T$ for pseudo-labels (Section 3.3 in the main paper).

In Table~\ref{table:ablation_projection_head}, we first show the benefit of having the projection head $h(\cdot)$ for multi-modal embedding space. We observe a 2\% performance gain by adding the projection head $h(\cdot)$, which demonstrates the importance of using $h(\cdot)$ for multi-modal regularization described in Section 3.2 of the main paper.
Moreover, we provide another ablation study where we add the projection head $h(\cdot)$ in the cross-domain module, while having $h(\cdot)$ for the cross-modal module as in our final model. Adding $h(\cdot)$ shows slightly worse results than our final model. One reason is that this scheme has less influence on the features that are supposed to be aligned for performing action recognition.

In Table \ref{table:ablation_setting}, we provide experimental results when different feature alignment methods are used in either cross-modal or cross-domain learning. In general, using the proposed contrastive learning method in both modules obtains the best performance, which shows the importance of having a unified contrastive learning framework for cross-modal and cross-domain
learning.

% We provide additional the ablation study of the adversarial alignment and projection head. In the second row in Table \ref{table:ablation_setting}, we show the performance of adversarial alignment on both cross-modal and cross-domain. In the third row, we add adversarial alignment to our cross-modal contrastive learning.
% We also show the effect of the additional projection head in cross-domain alignment. We observe that our final model design obtains the best performance.
% which is consistent with the observation in~\cite{simclr}. 
\begin{table}[!h]
	\caption{
		Ablation study on hyper-parameters on Epic-Kitchens. In the second group, we fix $T = 0.8$, while in the third group, we fix $\lambda = 1.25$.
	}
% 	\vspace{1mm}
	\label{table:ablation_hyper_params}
	\small
	\centering
	\renewcommand{\arraystretch}{1.1}
	\setlength{\tabcolsep}{3pt}
	\begin{tabular}{lc}
		\toprule
		
		Setting &   Mean \\
        \midrule
        Source-only & 45.5  \\
        Ours ($\lambda=1.25, T=0.8$) &  51.0\\
        \midrule
	    Ours ($\lambda=1.0$) &  50.1\\
	    Ours ($\lambda=1.5$) &  49.5\\
	    \midrule
	    Ours ($T=0.9$) &  49.6\\
	    Ours ($T=0.6$) &  49.8\\
		\bottomrule
	\end{tabular}
% 	\vspace{-3mm}
\end{table}

\begin{table}[!t]
	\caption{
		Ablation study on the projection head $h(\cdot)$ for EPIC-Kitchens.
	}
% 	\vspace{1mm}
	\label{table:ablation_projection_head}
	\small
	\centering
	\renewcommand{\arraystretch}{1.0}
	\setlength{\tabcolsep}{3pt}
	\resizebox{0.45\textwidth}{!}{
	\begin{tabular}{lcc}
		\toprule
		Setting &  $h(\cdot)$ in cross-modal module &   Mean \\
        \midrule
        Ours (modality) & \checkmark &  48.7\\
        Ours (modality) & \xmark &  46.7\\
        \midrule
        Setting &  $h(\cdot)$ in cross-domain module &   Mean \\
        \midrule
        Ours (modality + domain) & \checkmark &  50.1\\
        Ours-final (modality + domain) & \xmark &  51.0\\
		\bottomrule
	\end{tabular}}
% 	\vspace{-3mm}
\end{table}

\begin{table}[!t]
	\caption{
		Ablation study of different feature alignment methods on EPIC-Kitchens. ``Con.'' indicates our proposed contrastive learning approach, and ``Adv.'' denotes the adversarial learning scheme.
	}
% 	\vspace{1mm}
	\label{table:ablation_setting}
	\small
	\centering
	\renewcommand{\arraystretch}{1.1}
	\setlength{\tabcolsep}{3pt}
	\begin{tabular}{lllc}
		\toprule
		Setting & Modality & Domain &  Mean \\
		\midrule
        Con. (our final model) & Con. & Con. & 51.0 \\
        Adv. & Adv. & Adv. & 49.5 \\
        Adv. + Con. & Con. & Adv. & 50.1 \\
        % Our cross-modal contrastive learning & \multirow{2}{*}{50.1} \\
        % + adversarial cross-domain alignment & \\
        % Projection head in cross-domain alignment & 50.1 \\
		\bottomrule
	\end{tabular}
% 	\vspace{-3mm}
\end{table}

% \begin{table}[!t]
% 	\caption{
% 		Ablation study of the adversarial domain alignment and projection head.
% 	}
% % 	\vspace{1mm}
% 	\label{table:ablation_setting}
% 	\small
% 	\centering
% 	\renewcommand{\arraystretch}{1.1}
% 	\setlength{\tabcolsep}{3pt}
% 	\begin{tabular}{lc}
% 		\toprule
% 		Setting &   Mean \\
% 		\midrule
%         Ours (final) & 51.0 \\
%         Adv. across domains \& modalities & 49.5 \\
%         Our cross-modal contrastive learning & \multirow{2}{*}{50.1} \\
%         + adversarial cross-domain alignment & \\
%         % Projection head in cross-domain alignment & 50.1 \\
% 		\bottomrule
% 	\end{tabular}
% % 	\vspace{-3mm}
% \end{table}

\begin{figure*}[t]
	\centering
	\includegraphics[width=0.9\linewidth]{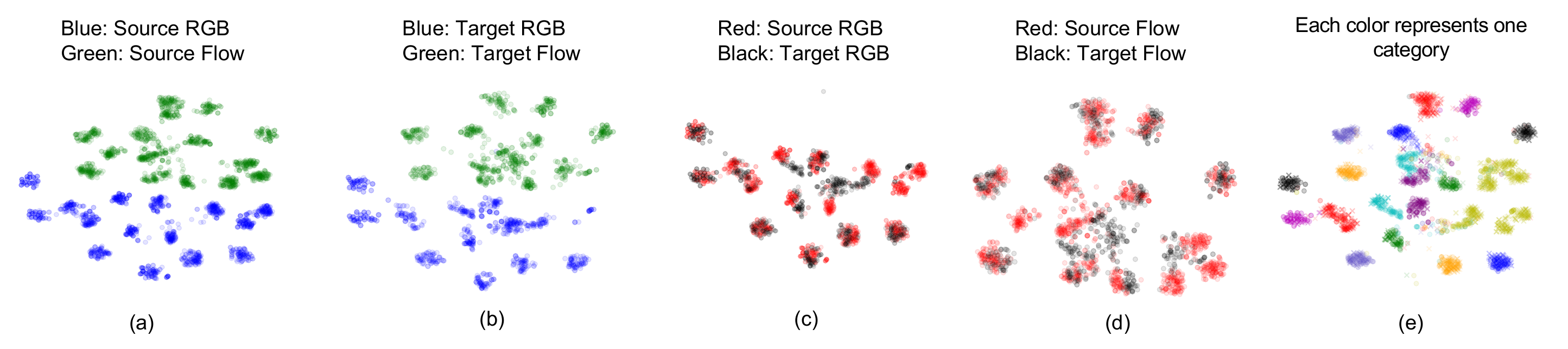}
	%\vspace{-4mm}
	\caption{
	t-SNE visualization on cross-modal and cross-domain features before the projection head $h(\cdot)$ on UCF $\rightarrow$ HMDB, \ie $F^a_s$, $F^m_s$, $F^a_t$, $F^m_t$. In (a)(b), we show the visualization for individual domains, where each domain contains the multi-modal features. In (c)(d), we visualize features for each modality, and each plot uses the features from two domains. (e) includes all the features from two domains and two modalities, where each color represents one action class.
	}
	\label{fig:tsne_supp}
	\vspace{-3mm}
\end{figure*}

\subsection{t-SNE Feature Visualizations}
Figure~\ref{fig:tsne_supp} shows different combinations of the
feature spaces before the projection head $h(\cdot)$, \ie, $F^a_s$, $F^m_s$, $F^a_t$, $F^m_t$. Figure~\ref{fig:tsne_supp}-(a,b) shows the RGB and flow features in each domain. While the RGB and flow features are almost completely aligned after the projection head in Figure 3 of the main paper, here the RGB and flow features still keep their respective information before the projection head $h(\cdot)$, which is useful for final action predictions.  Figures~\ref{fig:tsne_supp}-(c,d) shows how the source and target features are aligned in each modality, where our method can learn domain-invariant features.

\subsection{Visualizations for Cross-domain Retrievals}
\begin{figure*}[t]
% 	\begin{subfigure}{1.0\textwidth}
	\centering
	\includegraphics[width=0.9\linewidth]{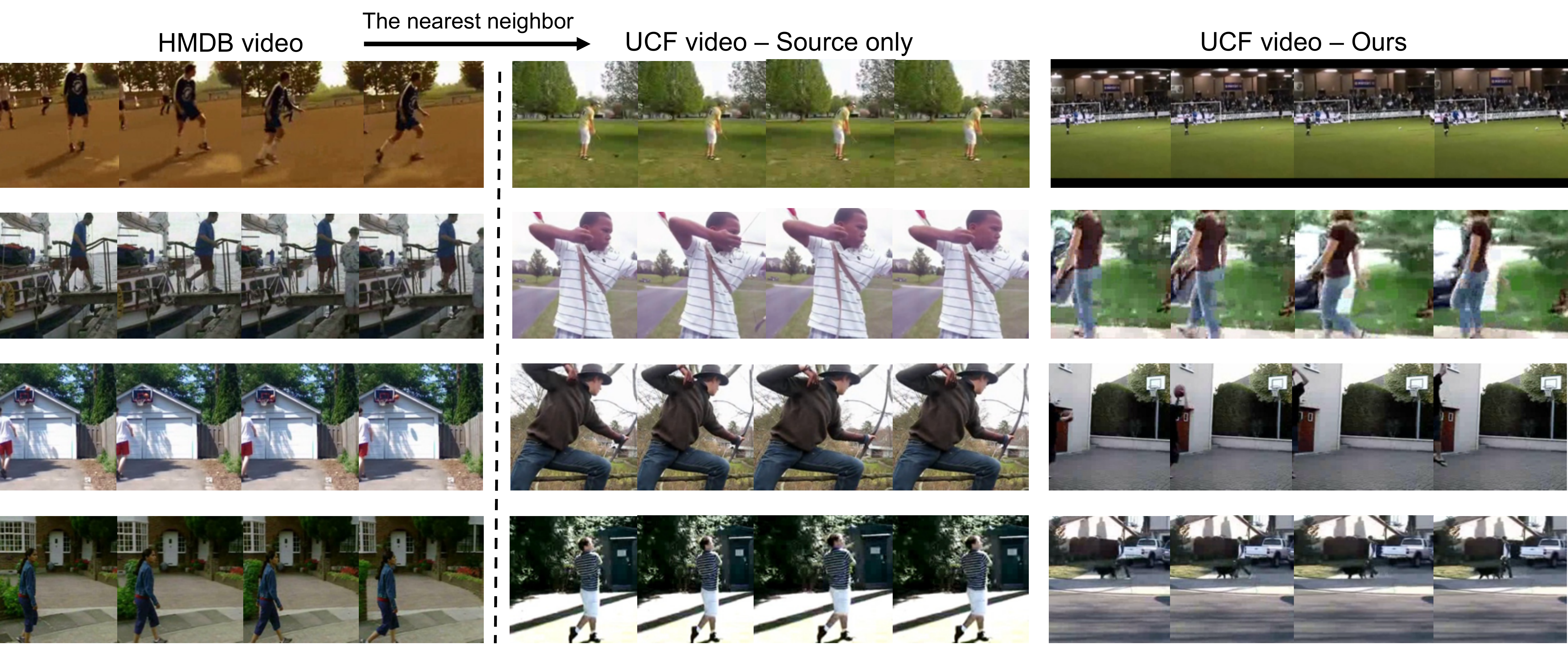} \\
	(a) UCF$\rightarrow$HMDB \\
% 	\caption{UCF$\rightarrow$HMDB}
% 	\end{subfigure}

    % \begin{subfigure}{1.0\textwidth}
    % \centering
    \includegraphics[width=0.9\linewidth]{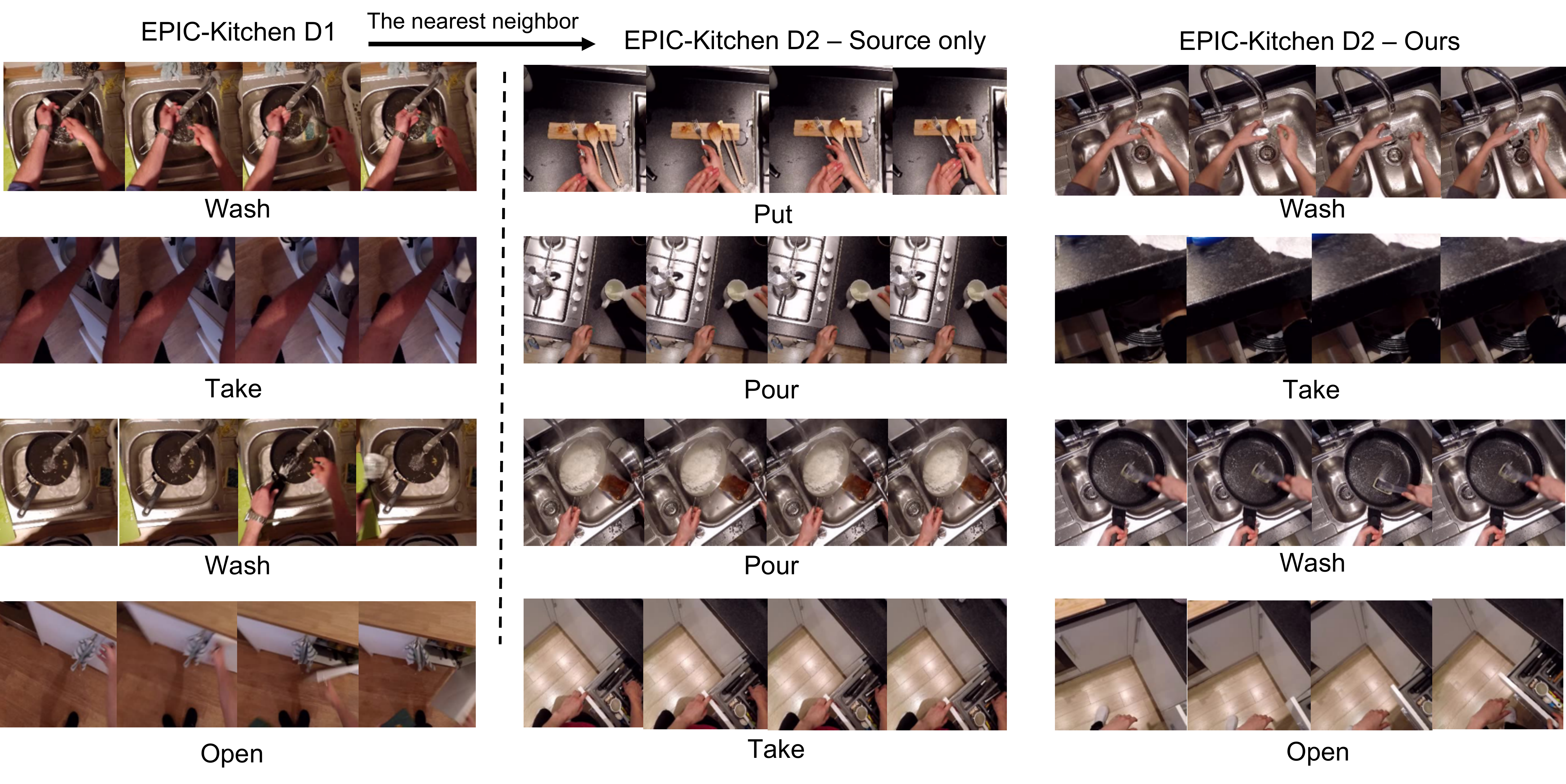} \\
    (b) EPIC-Kitchen D1$\rightarrow$D2
    % \caption{EPIC-Kitchen D1$\rightarrow$D2}
    % \end{subfigure}
    \vspace{2mm}
	\caption{Cross-domain retrievals in the RGB embedding space. Given the target feature $F_t^a$, we retrieve the closest neighbor $F_s^a$ in the source domain. By our contrastive learning framework, our model correctly aligns videos of the same class, while the source-only model are more likely to be biased to the background context.
	}
	\vspace{-4mm}
	\label{fig:retreival_supp}
\end{figure*}

In Figure~\ref{fig:retreival_supp}-(a), based on the target feature in HMDB, we show the nearest neighbor one from UCF. Similarly, we show the retrievals from EPIC Kitchens D1 and D2 in Figure~\ref{fig:retreival_supp}-(b). EPIC-Kitchen is more challenging than UCF-HMDB as it has more common background (\eg, similar kitchen backgrounds) or objects (\eg, frying pan, utensils) between different action classes. Our method shows better results that retrieve the videos of the same class.

\end{document}